\documentclass{article}

    \PassOptionsToPackage{numbers, compress}{natbib}
\usepackage[preprint]{neurips_2025}

\usepackage[utf8]{inputenc} % allow utf-8 input
\usepackage[T1]{fontenc}    % use 8-bit T1 fonts
\usepackage{hyperref}       % hyperlinks
\usepackage{url}            % simple URL typesetting
\usepackage{booktabs}       % professional-quality tables
\usepackage{amsfonts}       % blackboard math symbols
\usepackage{nicefrac}       % compact symbols for 1/2, etc.
\usepackage{microtype}      % microtypography
\usepackage{xcolor}         % colors
\usepackage{amsmath}
\usepackage{graphicx}
\usepackage{algorithmicx}
\usepackage{algorithm} % For algorithm environment
\usepackage{algpseudocode}
\usepackage{array}
\usepackage{booktabs}
\usepackage{pdflscape}
\usepackage{array}
\usepackage{multirow}
\usepackage{colortbl}
\usepackage{booktabs}
\usepackage{adjustbox}
\usepackage{tabularx}
\usepackage{pgfplots}
\usepackage{appendix}
\usepackage{subcaption}
\usepackage{hyperref}
\usepackage{enumitem}
\usepackage{titlesec}
\usepackage{colortbl} 
\usepackage{xfp}
\usepackage{tikz}
\usepackage[hypcap=false]{caption}

\usetikzlibrary{ arrows.meta, positioning}
\definecolor{color1}{HTML}{8DD3C7}
\definecolor{color2}{HTML}{fccde5}
\definecolor{color3}{HTML}{BEBADA}
\definecolor{color4}{HTML}{FB8072}
\definecolor{color5}{HTML}{80B1D3}
\definecolor{color6}{HTML}{FDB462}
\definecolor{color7}{HTML}{B3DE69}

\newcommand{\coloredcircle}[2][2.9]{%
  \raisebox{0pt}[0pt][0pt]{%
    \textcolor{gray!30}{\scalebox{#1}{\textbullet}}%
  }%
  \hspace{-2.2 ex}%
  \raisebox{.2 ex}{%
    \textcolor{#2}{\scalebox{\fpeval{.9*#1}}{\textbullet}}%
  }%
}

\titlespacing{\section}{0pt}{*0.5}{*0.5}     % {left}{before-sep}{after-sep}
\titlespacing{\subsection}{0pt}{*0.5}{*0.5}
\pgfplotsset{compat=1.18}
\title{Surrogate Interpretable Graph \\for Random Decision Forests}

\algnewcommand{\Continue}{\textbf{continue}}
\author{
    Akshat Dubey\\
    Center for Artificial Intelligence in Public Health Research (ZKI-PH)\\
    Robert Koch Institute\\
    Nordufer 20, 13353 Berlin \\
    Germany\\
    \texttt{DubeyA@rki.de} \\
    Department of Mathematics and Computer Science\\
    Free University of Berlin\\
    Arnimalle 14, 14195 Berlin\\
    Germany\\
    \texttt{dubea98@zedat.fu-berlin.de}
  \And
  Aleksandar Anźel \\
      Center for Artificial Intelligence in Public Health Research (ZKI-PH)\\
        Robert Koch Institute\\
        Nordufer 20, 13353 Berlin \\
        Germany\\
        \texttt{AnzelA@rki.de} \\
\And
    Georges Hattab\\
    Center for Artificial Intelligence in Public Health Research (ZKI-PH)\\
    Robert Koch Institute\\
    Nordufer 20, 13353 Berlin \\
    Germany\\
    Department of Mathematics and Computer Science\\
    Free University of Berlin\\
    Arnimalle 14, 14195 Berlin\\
    Germany\\
    \texttt{HattabG@rki.de} \\
    }
\begin{document}
\maketitle
\begin{abstract}
The field of health informatics has been profoundly influenced by the development of random forest models, which have led to significant advances in the interpretability of feature interactions. These models are characterized by their robustness to overfitting and parallelization, making them particularly useful in this domain. However, the increasing number of features and estimators in random forests can prevent domain experts from accurately interpreting global feature interactions, thereby compromising trust and regulatory compliance. A method called the surrogate interpretability graph has been developed to address this issue. It uses graphs and mixed-integer linear programming to analyze and visualize feature interactions. This improves their interpretability by visualizing the feature usage per decision-feature-interaction table and the most dominant hierarchical decision feature interactions for predictions. The implementation of a surrogate interpretable graph enhances global interpretability, which is critical for such a high-stakes domain. 
% The code is available at: \url{https://www.github.com/xxx}
\end{abstract}
\section{Introduction}

\noindent Random Forests (RF)~\cite{breiman2001random} or Random Decision Forests have become a cornerstone of health informatics~\cite{murris2024tree, li2020multicenter} and biomedicine~\cite{acharjee2020random} due to their unique ability to handle high-dimensional omics data while remaining robust to overfitting \cite{basu2018classification,boulesteix2012overview}. The model's ability to capture nonlinear relationships through hierarchical feature interactions~\cite{good2023feature} makes it particularly valuable for clinical decision support systems, where complex biomarker synergies often dictate patient outcomes \cite{hornung2021interaction,qi2011random}. However, as the number of features in the dataset grows, it becomes very important from an interpretability perspective to study feature interactions to enable the application of such methods to improve healthcare~\cite{stephan2015random, chew2022perceptions,potyka2023explaining}. Recent methodological advances attempt to bridge this interpretability gap using graphs, hierarchical shrinkage~\cite{agarwal2022hierarchical}, and rule mining. These include transforming RFs into bipolar argumentation frameworks, mapping tree decisions as arguments, identifying necessary feature sets, scaling into Markov networks, and using sampling and rule mining~\cite{potyka2023explaining, wang2023learning}. Rule-based interpretable random forest methods, such as SIRUS (Stable and Interpretable RUle Set)~\cite{benard2021sirus} and RAFREX (Random Forest Rule Extractor)~\cite{benard2021interpretable}, extract human-readable decision rules from complex tree ensembles, improving model transparency~\cite{benard2021interpretable}. However, the use of high-dimensional methods to analyze health data presents new scalability challenges, particularly for real-time decision support applications~\cite{kumar2024bioinformatics}. Current implementations struggle to balance comprehensive interaction analysis and linear time complexity~\cite{hornung2019tr, basu2018classification}. TreeSHAP~\cite{lundberg2018consistent, lundberg2020local}, a tree-based model, faces challenges in analyzing feature interactions, including exponential computational complexity, interaction blind spots, and path dependency artifacts. Advances such as TreeSHAP-IQ address these issues, but scalability problems remain in extreme scenarios, requiring dimensionality reduction or approximation techniques~\cite{muschalik2024beyond}.

\begin{figure}[htbp!]
\centering
\begin{subfigure}{0.5\textwidth}
  \centering
 \resizebox{0.75\linewidth}{!}{  
 \begin{tikzpicture}[
     node/.style={
         circle, draw=gray!30, line width=0.66pt, minimum size=.75cm, font=\large, text=black,
         inner sep=4pt, % padding inside nodes
         },
     edge/.style={
         ->, >=Stealth, thick, black, 
         shorten >=3pt, % gap before arrow tip
         shorten <=3pt  % gap after arrow tail (optional)
         },
     label/.style={
         font=\large, inner sep=1pt, fill=white, midway
         }
   ]

   % Node positions (manually adjusted for similarity to heptagonal shape)
 \node[node, fill=color1] (f1) at (3.00, 0.00) {};
 \node[node, fill=color2] (f2) at (1.87, 2.35) {};
 \node[node, fill=color3] (f3) at (-0.67, 2.92) {};
 \node[node, fill=color4] (f4) at (-2.70, 1.30) {};
 \node[node, fill=color5] (f5) at (-2.70, -1.30) {};
 \node[node, fill=color6] (f6) at (-0.67, -2.92) {};
 \node[node, fill=color7] (f7) at (1.87, -2.35) {};

 % Edges (matching your diagram)
 \draw[edge] (f1) -- node[label, above] {3} (f2);
 \draw[edge] (f1) -- node[label, right] {5} (f3);
 \draw[edge] (f1) -- node[label, right] {7} (f7);

 \draw[edge] (f2) -- node[label, above left] {2} (f4);

 \draw[edge] (f4) -- node[label, left] {4} (f3);
 \draw[edge] (f3) -- node[label, right] {2} (f7);
 % \draw[edge] (f7) -- node[label, right] {1} (f3);

 \draw[edge] (f4) -- node[label, left] {1} (f5);

 \draw[edge] (f5) -- node[label, below] {6} (f6);
 % \draw[edge] (f6) -- node[label, below] {6} (f5);

 \draw[edge] (f6) -- node[label, above] {2} (f7);

 % \draw[edge] (f7) -- node[label, below right] {3} (f1);

 \draw[edge] (f2) -- node[label, below right] {1} (f6);

 \end{tikzpicture}
 }
  \caption{Example Surrogate Interpretable Graph.}
  \label{fig:sub1}
\end{subfigure}%
\begin{subfigure}{0.4\textwidth}
  \centering
  \arrayrulecolor{white}
  \begin{tabular}{c|c|c|c|c|c|c|c}

 & 
 \coloredcircle[3]{color1} & 
 \coloredcircle[3]{color2} & 
 \coloredcircle[3]{color3} & 
 \coloredcircle[3]{color4} & 
 \coloredcircle[3]{color5} & 
 \coloredcircle[3]{color6} & 
 \coloredcircle[3]{color7} \\
\hline
$i_1$ & \cellcolor{gray!30} & \cellcolor{gray!30} &  &  &  & \cellcolor{gray!30} & \cellcolor{gray!30} \\ \hline
$i_2$ & \cellcolor{gray!30} & \cellcolor{gray!30} &  & \cellcolor{gray!30} & \cellcolor{gray!30} & \cellcolor{gray!30} & \cellcolor{gray!30} \\ \hline
$i_3$ & \cellcolor{gray!30} &  & \cellcolor{gray!30} &  &  &  & \cellcolor{gray!30} \\ \hline
$i_4$ & \cellcolor{gray!30} &  &  &  &  &  & \cellcolor{gray!30} \\ \hline
$i_5$ &  \cellcolor{gray!30} & \cellcolor{gray!30} & \cellcolor{gray!30} &  \cellcolor{gray!30} &  &  &  \cellcolor{gray!30} \\ \hline

  \end{tabular}
\arrayrulecolor{black}
  \vspace*{.66cm}

  \caption{Feature usage per Decision-Feature interaction.}
  \label{fig:sub2}
\end{subfigure}
\caption{Example Surrogate Interpretable Graph (SIG) and its feature usage per Decision-Feature interaction (DFI). 
(a) Example SIG visualized as an undirected graph with nodes arranged in a circle- with features visualized as the nodes and interaction frequencies as edge labels. Feature f1 is a originating feature for the hierarchical feature interaction due to the presence of all outward arrows, and feature f7 is the ending feature in the hierarchical feature interaction due to the presence of only inward arrows.
(b) Feature usage per decision-feature-interaction table for the example SIG, where connections are indicated as filled cells. 
Example interaction:
\textcolor{color1}{\textbf{Feature 1}}$\rightarrow$ \textcolor{color3}{\textbf{Feature 3}} $\rightarrow$ \textcolor{color7}{\textbf{Feature 7}} and 
\textcolor{color1}{\textbf{Feature 1}}$\rightarrow$
\textcolor{color2}{\textbf{Feature 2}}$\rightarrow$
\textcolor{color6}{\textbf{Feature 6}}$\rightarrow$
\textcolor{color7}{\textbf{Feature 7}}.}

% \coloredcircle[3]{color1}$\rightarrow$
% \coloredcircle[3]{color3}$\rightarrow$
% \coloredcircle[3]{color7}
% and
% \coloredcircle[3]{color1}$\rightarrow$
% \coloredcircle[3]{color2}$\rightarrow$
% \coloredcircle[3]{color6}$\rightarrow$
% \coloredcircle[3]{color7}

\label{fig:SIG}
\end{figure}
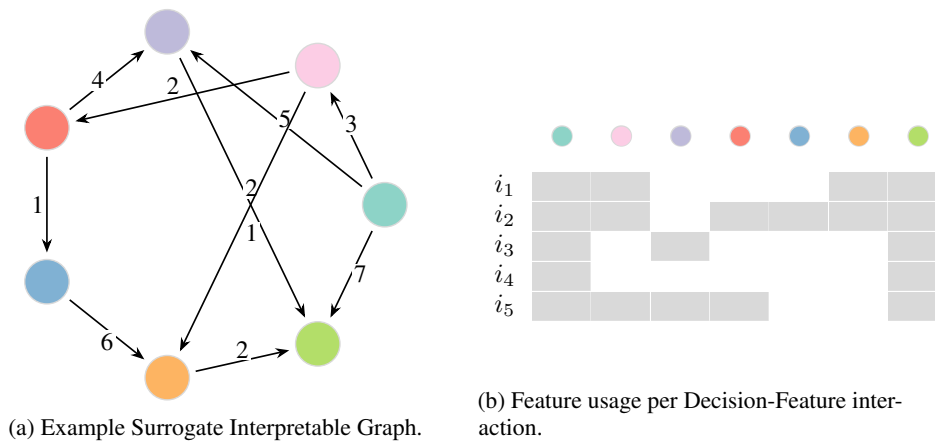

In this study, we have developed a framework for studying feature interactions using a surrogate interpretability graph (SIG). The goal is to capture global feature interactions, where nodes represent features and edges represent decision relationships learned by the model. The quantification of feature interactions is inferred from the frequency of co-occurring features within decision rules (Figure:~\ref{fig:sub1}). The approach combines SIG with Mixed Integer Linear Programming (MILP) to identify dominant decision patterns and optimize interpretability. It identifies redundant links and emphasizes the most salient feature relationships. The approach extracts salient features and presents a concise explanation of the model. A graph-based approach provides a visual and structured means of communicating the decision process, making it more trustworthy and user-oriented.
A table feature usage per decision-feature-interaction is also provided to facilitate the examination and interpretation of key feature interactions that contribute to the decision process(Figure:~\ref{fig:sub2}), which helps to address the AI rules for the high-stakes domain as well~\cite{dubey2024ai, dubey2024nested, ollech2020random}. Since TreeSHAP~\cite{bifet2022linear} is a widely used method for feature interaction analysis for tree methods, we have compared our method with the same. The study also shows that the time complexity of TreeSHAP increases with the \textit{f: }number of features and trees, leading to unbounded costs in the limit of f to infinity(Appendix:~\ref{appendix:maths}). In contrast, the SIG pipeline extracts rules once, vectorizes and clusters them, and constructs a feature-interaction graph that is optimized via MILP. The effective complexity remains lower in practical high-dimensional scenarios, regardless of the dataset size (Figure:~\ref{fig:sig_vs_treeshap_time}). Our method allows the study of hierarchical feature interactions underlying decision processes. Mathematical proofs are also provided to compare SIG with TreeSHAP's ability to analyze feature interactions. The proposed SIG framework has been implemented on four different datasets, each containing a different number of features and data points.
The proposed Surrogate Interpretable Graph method offers several key contributions:
\begin{itemize}
  \setlength\itemsep{0pt}
  \item Enhances model transparency by providing clear rule-based explanations for predictions,
  \item Improves user confidence by providing an intuitive visualization that shows the relationships and connections between features, features and attributes,  
  \item Achieves competitive performance compared to the state-of-the-art TreeSHAP method in terms of runtime complexity~\ref{appendix:comparison_treeshap_sig}, when the dataset has high number of features \& number of data points. This claim is also supported by mathematical proof~\ref{appendix:maths}.
\end{itemize}

% Further, we have derived and proved the asymptotic behavior of TreeSHAP, showing that its time complexity  
% \[
% \text{Time}_\text{TreeSHAP} = O\bigl(N \cdot T \cdot L^2 + N \cdot f^2\bigr)
% \]  
% grows prohibitively when the number of features \(f\) and the number of trees \(T\) increase—particularly due to the quadratic dependency on both the number of leaves \(L\) and features \(f\). Our mathematical proof confirmed that, in the limit \(f \to \infty\), TreeSHAP’s cost becomes unbounded even for moderate \(T\) and \(d\) (tree depth). In contrast, the SIG pipeline extracts rules once (\(O(T\cdot2^d)\)), vectorizes and clusters them (\(O(R\log R)\) for \(R\) rules), then constructs a feature‑interaction graph (\(O(R\cdot f^2)\)) which is subsequently optimized via MILP. In practical, high‑dimensional scenarios where each rule involves a small subset of features (\(k\ll f\)) and many rules are redundant (\(R'\ll R\)), the effective complexity  
% \[
% O\bigl(R\log R + R'\cdot k^2\bigr)
% \]  
% remains far lower than TreeSHAP’s, and becomes independent of the dataset size \(N\).
% We validated these findings with runtime experiments on four datasets of varying size and dimensionality, confirming that SIG+MILP matches or outperforms TreeSHAP when \(f\) or \(T\) grows beyond a few dozen (with speedups up to 10× in very high‑\(f\) regimes).

\section{Background}
RFs~\cite{breiman2001random} are ensemble learning methods that aggregate multiple decision trees to improve prediction accuracy and reduce overfitting. They play a critical role in healthcare by identifying complex biological mechanisms and advancing precision medicine, as they have the capability to completely reveal the datasets~\cite{ferry2024trained}. Iterative Random Forests (iRF)~\cite{basu2018iterative} detect higher-order interactions, providing insights into enhancer activity and alternative splicing. Generalized Random Forests (GRF)~\cite{athey2019generalized} model heterogeneous treatment effects, allowing clinicians to predict patient-specific responses to therapies. Methodological advances such as the Tree-based RFs feature importance and feature interaction network analysis framework (TBRFA)~\cite{yu2021deep} improve interpretability by disentangling feature importance and interaction networks. In nanotoxicology, RFs reveal how nanoparticle features interact to drive lung toxicity, highlighting the importance of interaction-aware models for risk assessment and safer nanomaterial design. TreeSHAP~\cite{lundberg2018consistent}, an efficient algorithm for computing SHAP~\cite{lundberg2017unified} values in tree-based models, has become a primary choice for studying feature interaction analysis~\cite{lundberg2018consistent}.
It uses the Shapley Interaction Index (SII), defined as:
\[
\text{SII}(S) = \sum_{T \subseteq \setminus S} \frac{(M - |T| - k)! |T|!}{(M - k + 1)!} \left( v(T \cup S) - v(T) \right),
\]
where 
\(S \) is a group of interacting features and $k = |S|$.
\( M \): Total number of input features in the model.
\( T \subseteq \setminus S \): A subset of features that excludes all elements in \( S \); i.e., \( T \cap S = \emptyset \).
 \( v(\cdot) \): The value function representing the model's output (e.g., prediction or expected value) when given a subset of features.
Natural language processing (NLP) represents text as a graph to capture semantic relationships and improve task performance. This approach uses graph theory to model connections between words, phrases, or entities, enabling advanced analysis using graph algorithms and machine learning techniques~\cite{probierz2023new}. Graphs transform feature interactions into intuitive networks, with nodes representing features and edges encoding the frequency or strength of their co-occurrence in decision paths~\cite{sirocchi2025feature}. Interaction forests extend this idea by performing explicit bivariate splits, constructing directed graphs with edge weights reflecting interaction strength and split hierarchy~\cite{hornung2022interaction, simon2023interpreting}. The Graph Random Forest (GRF) framework incorporates external domain knowledge, yielding subgraphs of highly connected features that improve accuracy and enhance biological interpretability~\cite{tian2023graph}. Rule extraction from ensemble tree-based models has been extensively studied. SIRUS extracts high-frequency rules by aggregating decision paths across bootstrap samples, while Random Forest Explainability (RFEX)~\cite{petkovic2018improving, petkovic2021random, rhodes2023geometry} compiles decision path rules from each tree into summary reports and ranks them by empirical support and conditional contribution. Interpretable trees (inTrees)~\cite{deng2019interpreting, aria2021comparison} prunes and filters raw rule sets to retain frequent, accurate patterns. Integer programming (IP) has emerged as a key technique for improving the interpretability of tree-based machine learning models by extracting concise rule sets from complex ensembles, preserving accuracy, and reducing complexity~\cite{moshkovitz2021connecting, goerigk2023framework, gunluk2021optimal, murali2023optimal}. Forest-ORE~\cite{haddouchi2025forest} formulates an optimal rule ensemble problem as a mixed-integer linear programming (MILP) problem, balancing predictability and interpretability to produce minimal yet high-coverage rule sets~\cite{chen2022representing, xu2021deep}. 
A Mixed-Integer Linear Programming (MILP) problem is formally defined as:
\begin{equation*}
\begin{array}{@{}c@{\hfill}c@{}}
\displaystyle
\min \sum_{(i,j) \in E} w_{ij} x_{ij}, 
&
\displaystyle
\quad\text{s.t.} \sum_{(i,j) \in \delta(S)} x_{ij} \geq 1, \quad \forall S \subset V, S \neq \emptyset, S \neq V, x_{ij} \in \{0,1\}, \forall (i,j) \in E
\\

\end{array}
\end{equation*}

% where \(x\_{ij}\): Binary decision variable indicating whether edge \((i,j)\) is selected, \(w\_{ij}\): Weight of edge \((i,j)\),and \delta\(S\): Set of edges with exactly one endpoint in subset \(S\) of vertices.\\
% \\
\begin{math}
x_{ij}: \text{Binary decision variable indicating whether edge } (i,j) \text{ is selected}\\
w_{ij}: \text{Weight of edge } (i,j) \\\delta(S): \text{Set of edges with exactly one endpoint in subset } S \text{ of vertices}.
\end{math}\\
MILP is a mixed-integer method for graph-based optimization problems that combines mathematics and interactions between continuous and discrete variables~\cite{ma2019mixed}. It represents problems as graphs with constraints and variables at separate vertices and edges, and combines traditional optimization with machine learning to improve solution efficiency. MILP supports graph matching, critical in computer vision and network alignment, and enforces network flow constraints for connectivity in path planning and network design. It also improves the visual clarity of graph layouts and generates well-distributed node placements in spatial optimization. Sparsification techniques further improve clarity by removing redundant edges without sacrificing functionality~\cite{ma2019mixed,li2024towards,liu2018scalable}.
\section{Surrogate Interpretable Graph Framework for Random Forest}
\subsection{Extracting decision rules from a random forest}
Decision paths can be extracted from RF, but they focus on how an instance is classified. Decision paths focus on a single data point at each instance classification and do not provide a global understanding of the individual trees. This results in a lack of insight into how features interact. By implementing the algorithm~\ref{alg:rf-rules}, we extract and store the rules by traversing each individual tree of the trained RF model. Extracting the decision rules from the individual trees allows us to capture the local decision-making, which we can study in the next steps of our methodology.
\begin{algorithm}[H]
\caption{Extraction of Decision Rules from Random Forest}
\label{alg:rf-rules}
\begin{algorithmic}[1]
\Require Trained Random Forest model $\mathcal{RF}$ with $T$ trees
\Ensure Set of decision rules $\mathcal{R}$
\State Initialize empty rule list $\mathcal{R}$
\For{each tree $t \in \mathcal{RF}$}
    \State Initialize stack $\mathcal{S}$ with root node and empty rule path
    \While{$\mathcal{S}$ is not empty}
        \State Pop $(node, rule\_path)$ from $\mathcal{S}$
        \If{$node$ is a leaf}
            \State Append rule "IF $rule\_path$ THEN class = $class$" to $\mathcal{R}$
        \Else
            \State Get $feature$ and $threshold$ from $node$
            \State Push right child with updated rule path $(feature > threshold)$
            \State Push left child with updated rule path $(feature \leq threshold)$
        \EndIf
    \EndWhile
\EndFor
\State \Return $\mathcal{R}$
\end{algorithmic}
\end{algorithm}
The decision rule extraction algorithm can be summarized as follows:\begin{itemize}[noitemsep]
\item The algorithm uses depth-first traversal with an explicit stack
\item Each rule is stored in $\mathcal{R}$ in the format: \\
\texttt{IF (feature1 $\leq$ x) AND (feature2 > y) THEN class = c}
\item Rules are constructed by concatenating conditions along the path using logical operations \textit{AND}.
\item The stack stores tuples of $(node, partial\_rule)$
\end{itemize}

\subsection{Standardizing and structuring rules for processing}
We need to standardize and structure the decision rule features in order to study the extracted rules. The features need to be named semantically, which adds complexity and inconsistency compared to the decision rules. The Scikit-Learn library's LabelEncoder converts feature names into numerical labels while preserving structure. Each feature is assigned a label and transformed into a standardized token (Algorithm:~\ref{alg:tfidf}). For example: \textit{Blood Pressure}$\,\to\,$\textit{FEAT\_3} and \textit{Glucose}$\,\to\,$\textit{FEAT\_7} 
\begin{algorithm}[H]
\caption{Rule Standardization and Tokenization}
\label{alg:standardize-tokenize}
\begin{algorithmic}[1]
\Require List of rules $R$, feature names $F$
\Ensure Tokenized and encoded rule list $T$
\State Initialize empty list $T$ and fit LabelEncoder $LE$ on $F$
\For{each rule $r \in R$}
    \State Split $r$ into components using ``AND''
    \State Initialize empty token list $tokens$
    \For{each component $c$}
        \If{$c$ contains ``$\leq$'' or ``$>$''}
            \State Extract $feature$ and $threshold$
            \State Encode $feature$ as $id \gets LE.\text{transform}(feature)$
            \State Append token in format ``FEAT\_id\_LTE\_threshold'' or ``FEAT\_id\_GT\_threshold''
        \EndIf
    \EndFor
    \State Join $tokens$ with ``AND'' and append to $T$
\EndFor
\State \Return $T$
\end{algorithmic}
\end{algorithm}
The example decision-rules and the corresponding encoding have been provided in Table:~\ref{tab:rule_encodings}.
\begin{table}[h]
    \centering
    \caption{Example of encoded representations corresponding to the decision rules.}
    \resizebox{\textwidth}{!}{%
    \begin{tabular}{cc}
        \toprule
       Encoded Representation  &  Decision Rules\\
       \midrule
        FEAT\_6\_LTE\_0.50) AND FEAT\_12\_LTE\_2.50) & (num\_major\_vessels <= 0.50) AND (thalassemia <= 2.50) \\
        % FEAT\_3\_LTE\_0.50) AND FEAT\_5\_GT\_93.00) & (exercise\_induced\_angina <= 0.50) AND (max\_heart\_rate\_achieved > 93.00) \\
        % \hline
        % FEAT\_5\_GT\_93.00) AND FEAT\_7\_GT\_139.00) &  (max\_heart\_rate\_achieved > 93.00) AND (resting\_blood\_pressure > 139.00) \\
        % \hline
        % FEAT\_10\_GT\_1.70) AND FEAT\_2\_GT\_267.00) & (st\_depression > 1.70) AND (cholesterol > 267.00) \\
        \bottomrule        
    \end{tabular}%
    }
    \label{tab:rule_encodings}
\end{table}
After conversion to an encoded representation, the decision rules are ready for tokenization because they're standardized. The encoded rules are more suitable for studying the decision rules than the semantic information. The tokenization step separates features, operators, and thresholds for further processing. Here's an example of a tokenized, coded decision rule: {(FEAT\_8\_LTE\_0.09) AND (FEAT\_26\_LTE\_18.64) AND (FEAT\_20\_GT\_957.45)}$\,\to\,${[ 'FEAT\_8', 'LTE', '0.09', 'AND', 'FEAT\_26', 'LTE', '18.64', 'AND', 'FEAT\_20', 'GT', '957.45']}. This process allows the rules to be structured, operators (LTE, GT) are preserved as tokens, thresholds (0.09, 18.64) are treated as part of the rule structure, and Boolean connectors (AND, OR) preserve logical dependencies.
\subsection{Rule representation via TF-IDF encoding}
After tokenizing the encoded rules, they are treated as text documents. Each rule is a sentence. For numerical representation, we use Term Frequency - Inverse Document Frequency (TF-IDF) (algorithm~\ref{alg:tfidf}). TF measures how often a token appears in a rule, while IDF assigns a score or weight to the frequency of the token across all rules (Table~\ref{tab:tfidf}). Features with higher TF-IDF scores contribute more to the decision. TF-IDF captures frequent patterns while downweighting rare ones.
\begin{table}
\caption{TF-IDF Matrix for two example rules and their assigned weights.}
\label{tab:tfidf}
\begin{tabular}{ccccccccccc}
\toprule
 Rule ID & FEAT\_8 & LTE & 0.09 & AND & LTE & FEAT\_26 & 18.64 & FEAT\_20 & GT & 957.45\\
 \midrule
 Rule 1 & 0.5 & 0.3 & 0.2 & 0.1 & 0.5 & 0.3 & 0.2 & 0.5 & 0.3 & 0.2 \\ 
 Rule 2 & 0.2 & 0.3 & 0.1 & 0.1 & 0.7 & 0.3 & 0.2 & 0.6 & 0.3 & 0.1 \\
\bottomrule
\end{tabular}
\end{table}
\begin{algorithm}[H]
\caption{TF-IDF Encoding of Decision Rules}
\label{alg:tfidf}
\begin{algorithmic}[1]
\Require Set of textual rules $\mathcal{R} = \{r_1, \ldots, r_n\}$
\Ensure TF-IDF matrix $\mathbf{M} \in \mathbb{R}^{n \times d}$
\State Tokenize each rule $r_i$ into terms $\mathcal{T}_i$; construct vocabulary $\mathcal{V} = \bigcup_{i=1}^n \mathcal{T}_i$ of size $d$
\For{each rule $r_i$ and term $t_j \in \mathcal{V}$}
    \State Compute term frequency: 
    $\mathrm{tf}_{i,j} = \frac{f_{i,j}}{\sum_{k=1}^d f_{i,k}}$
\EndFor
\For{each term $t_j \in \mathcal{V}$}
    \State Compute inverse document frequency: 
    $\mathrm{idf}_j = \log \frac{n}{1 + |\{r_i : t_j \in \mathcal{T}_i\}|}$
\EndFor
\State Compute TF-IDF matrix: 
$\mathbf{M}_{i,j} = \mathrm{tf}_{i,j} \cdot \mathrm{idf}_j$
\State Normalize each row: 
$\mathbf{M}_{i,:} \gets \frac{\mathbf{M}_{i,:}}{\|\mathbf{M}_{i,:}\|_2}$
\State \Return $\mathbf{M}$
\end{algorithmic}
\end{algorithm}
\subsection{Rule clustering using hierarchical clustering}
Extracting rules from RF can result in many redundant and overlapping rules. Clustering algorithms can reduce these rules by grouping similar rules together. This approach identifies common strategies by finding similar decision logic. It also reduces complexity and interpretation. Instead of a long list of rules, we use a small set of representative rules. This approach is superior to direct rule extraction. Without clustering, the dataset would have hundreds or thousands of redundant, difficult-to-navigate, and difficult-to-summarize rules. Clustering condenses these rules, preserving essential information while facilitating analysis. The coded rules are vectorized and clustered using their cosine similarity in an agglomerative manner (Algorithm:~\ref{alg:rule_clustering}).
The clustering strategy is as follows:
\begin{enumerate}[noitemsep]
    \item Cosine similarity measures the angular distance between the respective rule vectors. This allows for more robust clustering than Euclidean distance, which we found to be unsuitable for high-dimensional data.
    \item The further linking strategy merges the clusters based on their pairwise distances.
    The number of clusters should be chosen to achieve an approximate balance between overfitting and underfitting. According to our experiments, the optimal number of clusters should be the square root of f + N, rounded to the nearest perfect square.
\end{enumerate}
Let us define:
\begin{itemize}[noitemsep]
    \item $n$ = number of rules
    \item $d$ = dimensionality of the TF-IDF space
    \item Agglomerative Clustering is assumed to be from \textit{sklearn.cluster}
    \item labels\_ returns the assigned cluster for each rule.
\end{itemize}
\begin{algorithm}[H]
\caption{Rule Clustering via Agglomerative Clustering}
\label{alg:rule_clustering}
\begin{algorithmic}[1]
\Require TF-IDF matrix $M \in \mathbb{R}^{n \times d}$, number of clusters $k$
\Ensure Cluster labels $C \in \mathbb{Z}^n$
\State Initialize Agglomerative Clustering model:
\Statex \hspace{\algorithmicindent} $A \gets \text{AgglomerativeClustering}(n\_clusters = k)$
\State Fit the model to the TF-IDF matrix:
\Statex \hspace{\algorithmicindent} $A.\text{fit}(M)$
\State Retrieve cluster labels:
\Statex \hspace{\algorithmicindent} $C \gets A.\text{labels\_}$
\State \Return $C$
\end{algorithmic}
\end{algorithm}
Using this approach, we effectively group similar rules together. The overlapping decision paths are merged. This reduces redundancy. An example of clustering results is shown below:
\begin{enumerate}[noitemsep]
    \item Cluster 0:
      \begin{itemize}[noitemsep]
          \item FEAT\_6\_LTE\_0.50) AND FEAT\_12\_LTE\_2.50) AND FEAT\_3\_LTE\_0.50) AND FEAT\_5\_LTE\_93.00)
          \item FEAT\_6\_LTE\_0.50) AND FEAT\_12\_LTE\_2.50) AND FEAT\_3\_LTE\_0.50) AND FEAT\_5\_GT\_93.00) AND FEAT\_7\_LTE\_139.00) AND FEAT\_10\_LTE\_1.70)
      \end{itemize}
    \item Cluster 1:
      \begin{itemize}[noitemsep]
          \item FEAT\_10\_GT\_1.70) AND FEAT\_12\_LTE\_2.50) AND FEAT\_6\_GT\_0.50)
          \item FEAT\_10\_GT\_1.70) AND FEAT\_12\_GT\_2.50) AND FEAT\_1\_LTE\_2.50)
      \end{itemize}  
    % \item Cluster 2:
    %   \begin{itemize}[noitemsep]
    %       \item FEAT\_5\_LTE\_147.50) AND FEAT\_7\_LTE\_113.00) AND FEAT\_2\_LTE\_256.00) AND FEAT\_0\_LTE\_65.00)
    %       \item FEAT\_5\_LTE\_147.50) AND FEAT\_7\_LTE\_113.00) AND FEAT\_2\_LTE\_256.00) AND FEAT\_0\_GT\_65.00)
    %   \end{itemize}  
\end{enumerate}
\subsection{Creating surrogate interpretable graph (SIG)}
After the clustering process is complete, the feature interactions are transformed into a directed weighted graph (Algorithm:~\ref{alg:sig}). The graph is built as follows:
\begin{enumerate}[noitemsep]
    \item The feature transitions are extracted from the rules. 
    \begin{itemize}[noitemsep]
        \item (\textit{FEAT\_8}$\,\to\,$\textit{FEAT\_26})$\,\to\,$\textit{FEAT\_20}
    \end{itemize}
    \item The weight of the edge is equal to the frequency of co-occurrence in clustered rules.
    \item Implement MILP (Mixed-Integer Linear Programming) to optimize the sparsity of the graph.
    \begin{itemize}[noitemsep]
        \item The constraints are the number of edges \textit{k} to reduce the visual load while visualizing the surrogate graph.
        \item The goal is to preserve the connection with the most information.
    \end{itemize}
    \item The final optimized graph has the following properties:
    \begin{itemize}[noitemsep] \item The number of features is equal to the number of nodes.
        \item The number of edges are conditional transitions weighted by the frequency of co-occurrence.
        \item The features that are central to the predictions have high connectivity.
        \item The two-way arrows indicate that the connected features can occur in any order in the decision path. The one-way arrows indicate that the feature to which the arrow points can only occur if the feature from which the arrow originates occurs first.
    \end{itemize}
\end{enumerate}
\begin{algorithm}[H]
\caption{Construction of Surrogate Interpretable Graph (SIG) via MILP}
\label{alg:sig}
\begin{algorithmic}[1]
\Require Clustered rule sets $\{\mathcal{R}_1, \ldots, \mathcal{R}_k\}$, feature set $\mathcal{F}$
\Ensure Optimized Surrogate Interpretable Graph $G^*$
\State Initialize directed graph $G = (\mathcal{F}, E)$ with nodes $\mathcal{F}$ and empty edge set $E$
\For{each rule cluster $\mathcal{R}_i$}
    \For{each rule $r \in \mathcal{R}_i$}
        \For{each ordered feature pair $(f_s, f_t)$ in $r$}
            \State Add edge $(f_s \rightarrow f_t)$ to $E$
        \EndFor
    \EndFor
\EndFor
\State Define MILP variables $x_{uv} \in \{0,1\}$ for edge selection
\State Set objective: minimize $\sum x_{uv}$ (total selected edges)
\State Add constraints to:
\Statex \hspace{\algorithmicindent} $\bullet$ Preserve feature ordering from rules (path consistency)
\Statex \hspace{\algorithmicindent} $\bullet$ Ensure acyclic structure (DAG constraint)
\State Solve MILP to obtain optimal edge selections $x^*$
\State Construct $G^*$ using edges with $x^*_{uv} = 1$
\State \Return $G^*$
\end{algorithmic}
\end{algorithm}
\section{Results}
\noindent
\begin{minipage}[t]{0.45\textwidth}
    \centering
    \label{fig:sig_kidney_optimized}
    \includegraphics[width=\linewidth]{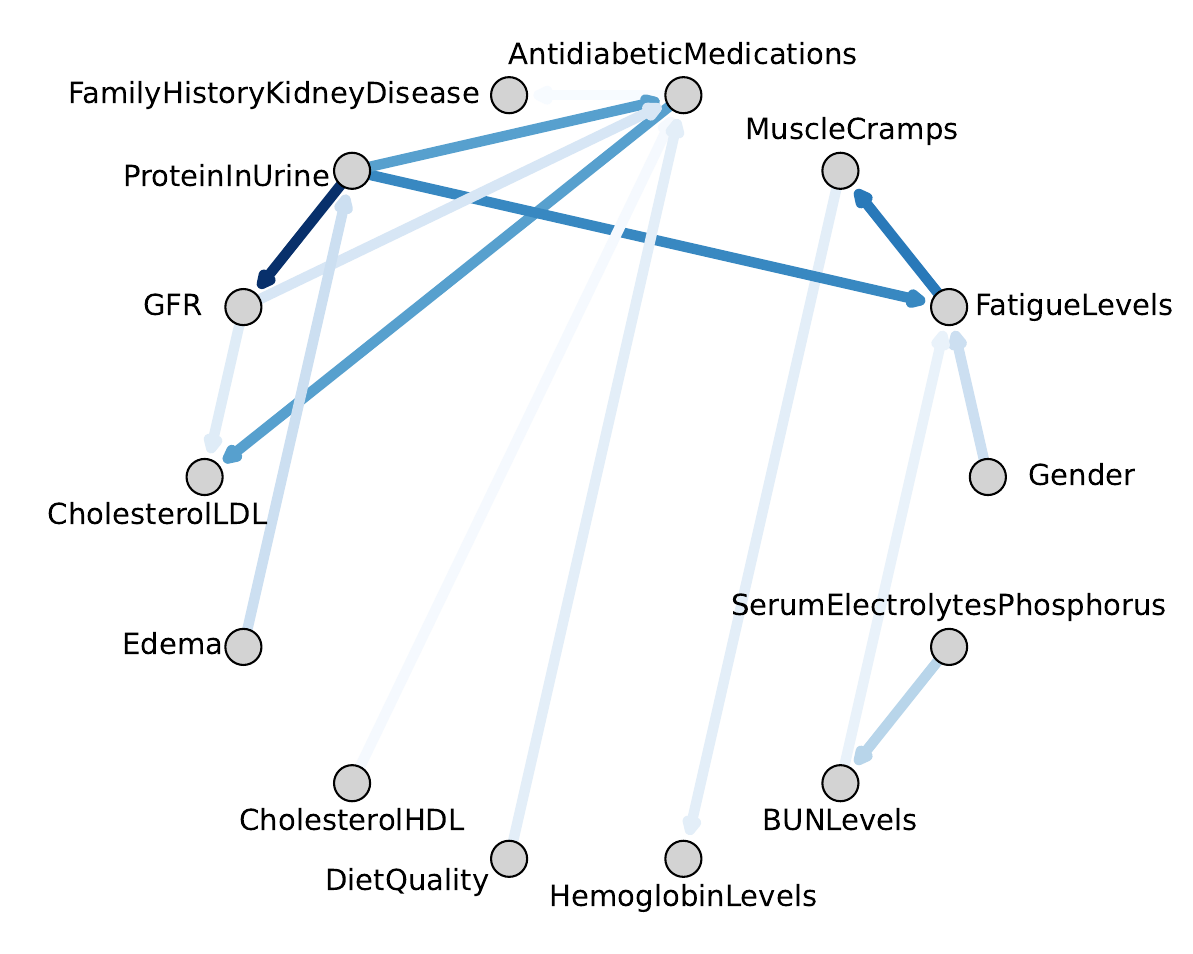}
    \captionof{figure}{The optimized surrogate interpretable graph (SIG) for the Chronic Kidney dataset. The SIG has been pruned to remove the nodes with minimal significance using MILP. The max number of edges for this optimized version was 15.}
\end{minipage}
\hfill
\begin{minipage}[t]{0.45\textwidth}
    \centering
    \label{fig:sig_aqi_optimized}
    \includegraphics[width=\linewidth]{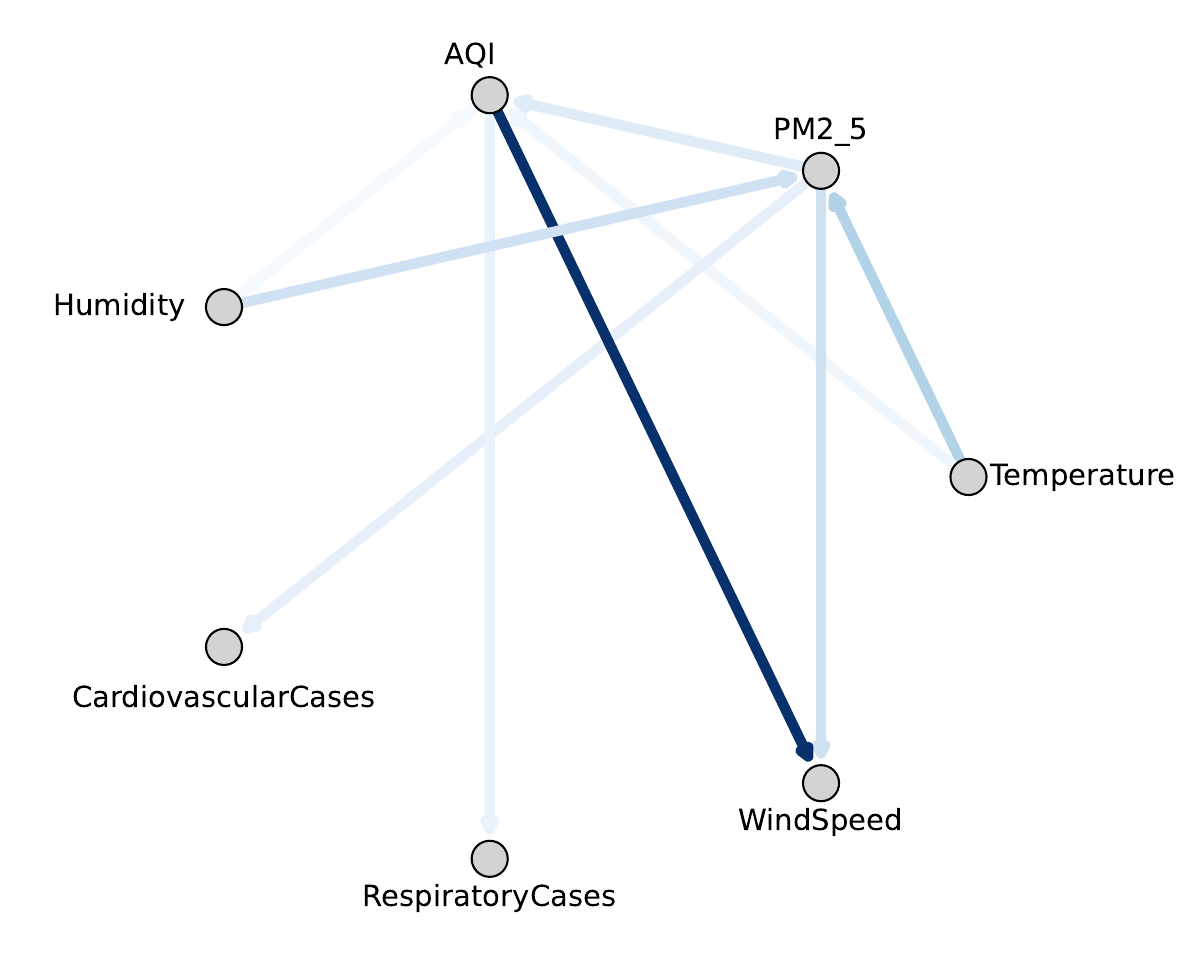}
    \captionof{figure}{The optimized surrogate interpretable graph (SIG) for the AQI and Health dataset. The SIG has been pruned to remove the nodes with minimal significance using MILP. The max number of edges for this optimized version was 15.}
\end{minipage}

The experimental procedures were performed on an Apple MacBook M1 Pro equipped with a 10-core central processing unit (CPU), a 16-core graphics processing unit (GPU), and 16 GB of random access memory (RAM). The RF classifier was evaluated and interpreted by partitioning the dataset into two subsets: a training set and a test set. The training set was used for model fitting. The test set was used for generalization performance and interpretability analysis. A fixed 80:20 split was used to maintain unbiasedness and representativeness of the data distribution. In cases of class imbalance, stratified sampling was used to maintain class proportions. This data partitioning strategy ensures reproducibility and prevents information leakage during interpretability evaluations. The SIG is generated for the trained model, allowing interpretation and retrieval of feature relationships. The SIG shows how individual trees work, allowing for the subset of features used by each tree in the RF model. The SIG also allows visualization and interpretation of feature co-occurrences, enhancing interpretability through a graph-based approach. Interpretations are communicated visually and structurally through a table for feature usage per DFI and hierarchical feature interactions. SIG allows interpretation of important DFIs, identifying dominant patterns for the global RF. We have also provided a detailed comparison between TreeSHAP and SIG (Appendix: ~\ref{appendix:comparison_treeshap_sig}). The analyses for the results section were performed on the chronic kidney disease (CKD)~\cite{kidney} and AQI and Health (AQIH)~\cite{aqi} datasets. Additional results for more datasets are available in the Appendix:~\ref{appendix:results}.
% \subsection{Chronic Kidney Diagnosis Dataset}
% This dataset~\cite{kidney} contains 54 features which include \textit{'PatientID', 'Age', 'Gender', 'Ethnicity', 'SocioeconomicStatus',
%        'EducationLevel', 'BMI', 'Smoking', 'AlcoholConsumption',
%        'PhysicalActivity', 'DietQuality', 'SleepQuality',
%        'FamilyHistoryKidneyDisease', 'FamilyHistoryHypertension',
%        'FamilyHistoryDiabetes', 'PreviousAcuteKidneyInjury',
%        'UrinaryTractInfections', 'SystolicBP', 'DiastolicBP',
%        'FastingBloodSugar', 'HbA1c', 'SerumCreatinine', 'BUNLevels', 'GFR',
%        'ProteinInUrine', 'ACR', 'SerumElectrolytesSodium',
%        'SerumElectrolytesPotassium', 'SerumElectrolytesCalcium',
%        'SerumElectrolytesPhosphorus', 'HemoglobinLevels', 'CholesterolTotal',
%        'CholesterolLDL', 'CholesterolHDL', 'CholesterolTriglycerides',
%        'ACEInhibitors', 'Diuretics', 'NSAIDsUse', 'Statins',
%        'AntidiabeticMedications', 'Edema', 'FatigueLevels', 'NauseaVomiting',
%        'MuscleCramps', 'Itching', 'QualityOfLifeScore', 'HeavyMetalsExposure',
%        'OccupationalExposureChemicals', 'WaterQuality',
%        'MedicalCheckupsFrequency', 'MedicationAdherence', 'HealthLiteracy',
%        'Diagnosis', 'DoctorInCharge'}.
The CKD patient dataset includes demographic details, lifestyle factors, medical history, clinical measurements, medication use, symptoms, quality of life scores, environmental exposures, and health behaviors. The trained RF model is used to generate the feature interaction graph. The graph is optimized \& pruned using MILP to retain the most important features with the highest interaction and remove the features with minimal interaction \& contribution. The result is a SIG. After optimization, the resulting features are \textit{f1: AntidiabeticMedications, f2: MuscleCramps, f3: FatigueLevels, f4: Gender, f5: SerumElectrolytesPhosphorus, f6: BUNLevels, f7: HemoglobinLevels, f8: DietQuality, f9: CholestrolHDL, f10: Edema, f11: CholestrolLDL, f12: GFR, f13: ProteinUrine, f14: FamilyHistoryKidneyDisease}. From the SIG (Figure:~\ref{fig:sig_kidney_optimized}), we can conclude that there are 10 major DFIs that are involved in the decision most of the time. The feature \textit{f1} appears in 6 DFIs. \textit{f2} appears in 3 DFIs. \textit{f3} appears in 3 DFIs. \textit{f4} occurs in 1 DFIs. \textit{f5} occurs in 1 DFIs. \textit{f6} occurs in 1 DFIs. \textit{f7} occurs in 2 DFIs. \textit{f8} is encountered in 2 DFIs. \textit{f9} is encountered in 2 DFIs. \textit{f10} is found in 4 DFIs. \textit{f11} is found in 4 DFIs.\textit{f12} is found in 4 DFIs. \textit{f13} is found in 4 DFIs. \textit{f14} is found in 3 DFIs. The features responsible for the decision are \textit{f5, f4, f8, f9, f10}. DFI 6 and DFI 10 are the feature interactions with the most features (Table:~\ref{tab:adjacency_matrix_sig_kidney}). The table shows the hierarchical feature interaction for the features used by the most single trees in the RF classifier. Alternatively, it can be interpreted as the order of the features that appear in the most decision paths. The number of unique DFIs is 10, which also shows that the number of estimators chosen for the RF classifier, i.e., ~15, is optimal.

\noindent
\begin{minipage}[t]{0.47\textwidth}
\centering
\captionof{table}{Feature usage per decision-feature-interaction (DFI) from SIG for the Chronic Kidney dataset (Figure:~\ref{fig:sig_kidney_optimized}).}
\label{tab:adjacency_matrix_sig_kidney}
\begin{adjustbox}{width=\textwidth}
\renewcommand{\arraystretch}{1.87}
\arrayrulecolor{white}
\begin{tabular}{c|c|c|c|c|c|c|c|c|c|c|c|c|c|c}
%\toprule
\textbf{DFI} & \textbf{f1} & \textbf{f2} & \textbf{f3} & \textbf{f4} & \textbf{f5} & \textbf{f6} & \textbf{f7} & \textbf{f8} & \textbf{f9} & \textbf{f10} & \textbf{f11} & \textbf{f12} & \textbf{f13} & \textbf{f14} \\ \hline
%\midrule
\(i_1\) & \cellcolor{gray!30}&  &  &  &  &  &  & \cellcolor{gray!30}&  &  &  &  &  & \cellcolor{gray!30}\\ \hline
\(i_2\) & \cellcolor{gray!30}&  &  &  &  &  &  & \cellcolor{gray!30}&  &  & \cellcolor{gray!30}&  &  &  \\ \hline
\(i_3\) & \cellcolor{gray!30}&  &  &  &  &  &  &  & \cellcolor{gray!30}&  &  &  &  & \cellcolor{gray!30}\\ \hline
\(i_4\) & \cellcolor{gray!30}&  &  &  &  &  &  &  & \cellcolor{gray!30}&  & \cellcolor{gray!30}&  &  &  \\ \hline
\(i_5\) &  &  &  &  &  &  &  &  &  & \cellcolor{gray!30}& \cellcolor{gray!30}& \cellcolor{gray!30}& \cellcolor{gray!30}&  \\ \hline
\(i_6\) & \cellcolor{gray!30}&  &  &  &  &  &  &  &  & \cellcolor{gray!30}&  & \cellcolor{gray!30}& \cellcolor{gray!30}& \cellcolor{gray!30}\\ \hline 
\(i_7\) & \cellcolor{gray!30}&  &  &  &  &  &  &  &  & \cellcolor{gray!30}& \cellcolor{gray!30}& \cellcolor{gray!30}& \cellcolor{gray!30}&  \\ \hline
\(i_8\) &  & \cellcolor{gray!30}& \cellcolor{gray!30}&  &  &  & \cellcolor{gray!30}&  &  & \cellcolor{gray!30}&  & \cellcolor{gray!30}& \cellcolor{gray!30}&  \\ \hline
\(i_9\) &  & \cellcolor{gray!30}& \cellcolor{gray!30}& \cellcolor{gray!30}&  &  &  &  &  &  &  &  &  &  \\ \hline
\(i_{10}\) &  & \cellcolor{gray!30}& \cellcolor{gray!30}&  & \cellcolor{gray!30}& \cellcolor{gray!30}& \cellcolor{gray!30}&  &  &  &  &  &  &  \\
%\bottomrule
\end{tabular}
\arrayrulecolor{black}
\end{adjustbox}
\end{minipage}%
\hfill
\begin{minipage}[t]{0.45\textwidth}
\centering
\captionof{table}{Hierarchical feature interaction for the features with most interactions for the Chronic Kidney dataset (Figure:~\ref{fig:sig_kidney_optimized}).}
\label{tab:decision_path_per_tree_sig_kidney}
\begin{adjustbox}{width=\textwidth}
\begin{tabular}{cl}
%\toprule
\textbf{DFI} & \textbf{Hierarchical Feature Interaction} \\
%\midrule
\(i_1\) & f8 $\rightarrow$ f1 $\rightarrow$ f14 \\
\(i_2\) & f8 $\rightarrow$ f1 $\rightarrow$ f11 \\
\(i_3\) & f9 $\rightarrow$ f1 $\rightarrow$ f14 \\
\(i_4\) & f9 $\rightarrow$ f1 $\rightarrow$ f11 \\
\(i_5\) & f10 $\rightarrow$ f13 $\rightarrow$ f12 $\rightarrow$ f11 \\
\(i_6\) & f10 $\rightarrow$ f13 $\rightarrow$ f12 $\rightarrow$ f1 $\rightarrow$ f14 \\
\(i_7\) & f10 $\rightarrow$ f13 $\rightarrow$ f12 $\rightarrow$ f1 $\rightarrow$ f11 \\
\(i_8\) & f10 $\rightarrow$ f13 $\rightarrow$ f3 $\rightarrow$ f2 $\rightarrow$ f7 \\
\(i_9\) & f4 $\rightarrow$ f3 $\rightarrow$ f2 $\rightarrow$ f7 \\
\(i_{10}\) & f5 $\rightarrow$ f6 $\rightarrow$ f3 $\rightarrow$ f2 $\rightarrow$ f7 \\
%\bottomrule
\end{tabular}
\end{adjustbox}
\end{minipage}%
% \subsection{AQI and Health Dataset}

% This dataset~\cite{aqi} contains 15 features \textit{'AQI', 'PM10', 'PM2\_5', 'NO2', 'SO2', 'O3', 'Temperature', 'Humidity', 'WindSpeed', 'RespiratoryCases', 'CardiovascularCases', 'HospitalAdmissions', 'HealthImpactScore', 'HealthImpactClass', 'ReportID'}. The feature \textit{ReportID} was dropped. 
% For detailed description of the features, please refer to Appendix:~\ref{appendix:aqi_data}.
The AQIH dataset contains comprehensive information on air quality and its impact on public health for 5,811 datasets. The study includes variables such as the Air Quality Index (AQI), pollutant concentrations, meteorological conditions, and health impact metrics. The target variable is the health impact class, which categorizes health impacts based on air quality and related factors. The trained RF model is used to generate a graphical representation.

\noindent
\begin{minipage}{0.48\textwidth}
\centering
\captionof{table}{Feature usage per decision-feature-interaction (DFI) for the AQI and Health dataset (Figure:~\ref{fig:sig_aqi_optimized}).}
\label{tab:adjacency_matrix_aqi_heart}
\arrayrulecolor{white} % set vertical and horizontal lines color to white
\begin{adjustbox}{width=\textwidth}
\begin{tabular}{c|c|c|c|c|c|c|c}
\toprule
\textbf{DFI} & \textbf{f1} & \textbf{f2} & \textbf{f3} & \textbf{f4} & \textbf{f5} & \textbf{f6} & \textbf{f7} \\
\midrule
\(i_1\) &  & \cellcolor{gray!30}& \cellcolor{gray!30}& \cellcolor{gray!30}&  &  &  \\ \hline
\(i_2\) &  & \cellcolor{gray!30}& \cellcolor{gray!30}&  &  & \cellcolor{gray!30}&  \\ \hline
\(i_3\) & \cellcolor{gray!30}&  & \cellcolor{gray!30}& \cellcolor{gray!30}&  &  &  \\ \hline
\(i_4\) & \cellcolor{gray!30}&  & \cellcolor{gray!30}&  & \cellcolor{gray!30}&  &  \\ \hline
\(i_5\) &  & \cellcolor{gray!30}& \cellcolor{gray!30}& \cellcolor{gray!30}&  &  & \cellcolor{gray!30}\\ \hline
\(i_6\) &  & \cellcolor{gray!30}& \cellcolor{gray!30}&  &  & \cellcolor{gray!30}& \cellcolor{gray!30}\\
\bottomrule
\end{tabular}
\arrayrulecolor{black} % reset color if needed afterwards
\end{adjustbox}
\end{minipage}%
\hfill
\begin{minipage}{0.48\textwidth}
\centering
\captionof{table}{Hierarchical feature interaction for features with most number of interactions for the AQI and Health dataset (Figure:~\ref{fig:sig_aqi_optimized}).}
\label{tab:decision_path_per_tree_sig_aqi}
\begin{adjustbox}{width=\textwidth}
\renewcommand{\arraystretch}{1.35}
\begin{tabular}{cp{6.2cm}}
%\toprule
\textbf{DFI} & \textbf{Hierarchical Feature Interaction} \\
%\midrule
\(i_1\) & f3 $\rightarrow$ f2 $\rightarrow$ f4 \\
\(i_2\) & f3 $\rightarrow$ f2 $\rightarrow$ f6 \\
\(i_3\) & f3 $\rightarrow$ f1 $\rightarrow$ f4 \\
\(i_4\) & f3 $\rightarrow$ f1 $\rightarrow$ f5 \\
\(i_5\) & f3 $\rightarrow$ f7 $\rightarrow$ f2 $\rightarrow$ f4 \\
\(i_6\) & f3 $\rightarrow$ f7 $\rightarrow$ f2 $\rightarrow$ f6 \\
%\bottomrule
\end{tabular}
\end{adjustbox}
\end{minipage}
\renewcommand{\arraystretch}{1.3}
\noindent

The graph is optimized and pruned using a mixed-integer linear programming (MILP) method to retain the most important features with the highest interaction and remove the features with minimal interaction or contribution. This results in a spatial interpologram (SIG; Figure~\ref{fig:sig_aqi_optimized}). After optimization, the resulting features are f1 (AQI), f2 (PM2.5), f3 (temperature), f4 (wind speed), f5 (respiratory cases), f6 (cardiovascular cases), and f7 (humidity).
We can interpret from the SIG that there exist 6 major DFIs (Table:~\ref{tab:adjacency_matrix_aqi_heart}) which ontribute to the decision-making majorly. The \textit{f1} is encountered in 2 DFIs. The \textit{f2} is encountered in 4 DFIs. The \textit{f3} is encountered in 6 DFIs. The \textit{f4} is encountered in 3 DFIs. The \textit{f5} is encountered in 1 DFIs. The \textit{f6} is encountered in 2 DFIs. The \textit{f7} is encountered in 7 DFIs. The most important feature which is encountered in all the DFIs is \textit{f3}. From the hierarchical feature interaction (Table:~\ref{tab:decision_path_per_tree_sig_aqi}).
\section{Conclusion}
In this work, we presented the Surrogate Interpretable Graph (SIG) combined with MILP for feature interaction analysis in RF.  We also showed how the SIG can be summarized as a table for feature usage per DFI that provides interpretations of the most important decision-feature interactions (DFIs) that contribute to decision making for the features they use, and how hierarchical feature interactions occur. Finally, focusing on healthcare domains characterized by a high number of molecular or clinical features, we show that SIG+MILP scales gracefully and provides clear, global explanations where TreeSHAP becomes infeasible. We have provided a comparison table between SIG and TreeSHAP to highlight the differences and ease of use~\ref{tab:sig_vs_tree_shap}. Overall, while TreeSHAP remains a gold standard for local, instance-level attributions, its quadratic blow-up in features limits its use for global, high-dimensional interpretability. In contrast, SIG+MILP provides a compressed and global surrogate that directly reflects the tree logic, making it particularly well-suited for large-scale applications.

\begin{table}[h!]
\centering
\caption{Comparison of Surrogate Interpretable Graph and TreeSHAP Interaction Values}
\label{tab:shap_interaction_vs_sig}
\begin{adjustbox}{width=0.99\textwidth}
\begin{tabularx}{\textwidth}{lXX}
\toprule
\textbf{Method} & \textbf{Surrogate Interpretable Graph (SIG)} & \textbf{SHAP Interaction Values} \\
\midrule
\textbf{Transparency} & High – Shows actual rules/features the model uses. & Medium – SHAP values are additive but not always easily explainable. \\

\textbf{Structure Awareness} & Preserves feature hierarchy and paths from trees. & Abstracts away from the actual model structure. \\

\textbf{Global Interpretability} & Excellent – Shows average co-occurrence and rule strength across forest. & Weaker – Harder to globally interpret interactions beyond pairwise SHAP plots. \\

\textbf{Rule-Based} & Yes – Derives and clusters real decision rules. & No – Outputs contributions for a given prediction. \\

\textbf{Visual Simplicity} & High–Graph is pruned and optimized. & SHAP plots (force, dependence) can be overwhelming. \\

\textbf{Faithfulness to Trees} & Very faithful – Graph and rules are extracted directly from trees. & Approximate – SHAP relies on game theory assumptions. \\

\textbf{Customizability} & High – Graph structure, clustering, rule depth, etc., can be adjusted. & Limited – SHAP has a fixed format and logic. \\
\bottomrule
\end{tabularx}
\end{adjustbox}
\end{table}

\section{Limitations and Future Works}
The SIG has demonstrated satisfactory functionality; however, its impact on various MILP solvers~\cite{geng2023deep} requires further investigation. The study is not without limitations, including the lack of real-world domain expert user studies. The potential of boosting models such as XGBoost~\cite{chen2016xgboost} for further investigation is promising. The approach provides interpretability at a global level, which is a crucial aspect in high-stakes domains. In subsequent studies, the strengths of TreeSHAP's local attributions will be integrated with SIG's global rule graph, allowing for instance-specific and cohort-level insights. Incremental or streaming MILP solvers are essential for preserving interpretability in dynamic settings. Adapting SIG with domain knowledge ensures that feature interactions respect known molecular hierarchies~\cite{zhang2023general}, thereby enhancing clinical confidence and actionable hypotheses.
% In the context of single-cell genomics and high-throughput screening, the utilization of approximate or distributed algorithms, such as locality-sensitive hashing, is imperative for the effective management of rule sets. 

% % TreeSHAP visualizations in real diagnostic workflows. Integration with Model Development will be crucial factor to 
% % embed surrogate graph construction into automated machine learning pipelines, enabling automated selection of interpretable models and real‑time interpretability auditing during model training. By pursuing these directions, we aim to bridge the gap between robust and practical interpretability in the most demanding, high‑dimensional arenas of healthcare and beyond.

\clearpage 
\bibliographystyle{ieeetr}
\bibliography{References}

\clearpage
\appendix
\section{Mathematical Proof}
\label{appendix:maths}
We have:
\begin{flalign}
    f &= \text{number of features} \\
    T &= \text{number of trees} \\
    d &= \text{maximum depth of trees} \\
    L &\approx 2^d = \text{number of leaves per tree} \\
    R &= \text{number of extracted rules} = O(T \cdot L) \\
    N &= \text{number of data instances}
\end{flalign}

We assume:
\begin{itemize}
    \item $f \to \infty$ (feature explosion)
    \item $T \to \infty$ (dense forest)
    \item Goal: compute all pairwise interactions for accurate local attribution
\end{itemize}

\subsection*{TreeSHAP for Feature Interaction Analysis}

\textbf{TreeSHAP Time Complexity:}
\[
\text{Time}_{\text{TreeSHAP}} = O(N \cdot T \cdot L^2 + N \cdot f^2)
\]

\textbf{Why TreeSHAP is Used:}
\begin{itemize}
    \item Captures both main and interaction effects among features
    \item Attribution is computed per instance, offering local interpretability
    \item Efficient implementation using recursive decomposition and dynamic programming
\end{itemize}

\textbf{Limitations:}
\begin{itemize}
    \item Each instance requires a full SHAP interaction matrix: $f^2$ values
    \item Grows quadratically with the number of leaves per tree ($L^2$)
    \item Linearly dependent on the number of data points $N$
\end{itemize}

\subsection*{Comparison When $f \rightarrow \infty$ and $T$ Grows}

Assume:
\begin{itemize}
    \item $L = 2^d$ (balanced binary trees)
    \item $k = o(f)$: average number of meaningful features per SHAP interaction
\end{itemize}

\[
\text{Time}_{\text{TreeSHAP}} = O(N \cdot T \cdot L^2 + N \cdot f^2)
\]

As $f$ grows large, the SHAP interaction matrix scales as $f^2$ per instance. The tree traversal cost also increases rapidly with $T$ and $L$.

\subsection*{Formal Asymptotic Behavior}

\[
\lim_{f \to \infty} \text{Time}_{\text{TreeSHAP}} = \infty \quad \text{(quadratic in features and leaves)}
\]

Hence, for high-dimensional settings or deep forests, TreeSHAP interaction analysis becomes computationally expensive.

\subsection*{Time Complexity Assumptions and Comparative Growth}

Let us define the following variables:

\vspace{1em}
\noindent \textbf{Assumptions}:
\begin{itemize}
    \item Feature space explosion: $f \to \infty$
    \item Dense forest assumption: $T \to \infty$
    \item Moderate depth per tree: $d$ is constant
    \item Goal: Perform pairwise interaction-based attribution using TreeSHAP
\end{itemize}

\subsection*{TreeSHAP with Feature Interaction Values}
\begin{itemize}
    \item Computes $O(f^2)$ SHAP interaction values per instance
    \item Tree traversal time: $O(T \cdot L^2)$
    \item Overall time complexity:
    \[
        \text{Time}_{\text{TreeSHAP}} = O(N \cdot T \cdot L^2 + N \cdot f^2) \quad 
    \]
\end{itemize}

\subsection*{Scalability Implications}
\begin{itemize}
    \item For small $f$, TreeSHAP gives accurate and efficient feature interactions
    \item For large $f$ or $T$, the computational burden grows sharply
    \item Memory and time cost for full SHAP interaction tensor becomes a bottleneck
\end{itemize}

\section{Additional Results}
\label{appendix:results}
% \begin{figure}[htbp]

%     \caption{Optimized surrogate interpretable graphs (SIGs) across additional datasets. The graphs have been
% pruned using MILP to retain only significant nodes}

\begin{figure}
    % \centering
    \includegraphics[width=\linewidth]{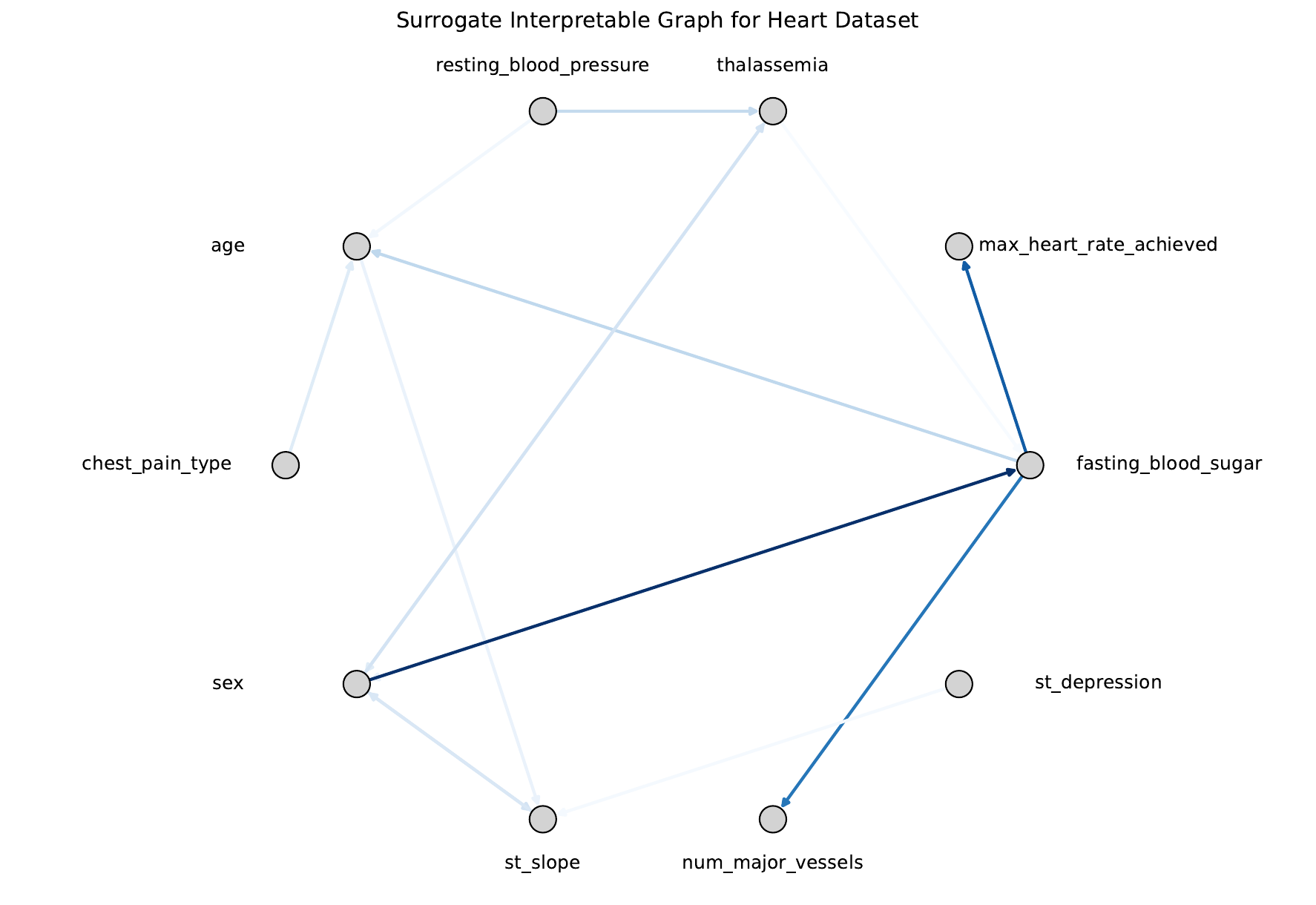}
    \caption{The optimized surrogate interpretable graph (SIG) for the heart dataset. The SIG has been pruned to remove the nodes with minimal significance using MILP. The max number of edges for this optimized version was 15.}
    \label{fig:sig_heart_optimized}
\end{figure}

\begin{figure}
        \includegraphics[width=\linewidth]{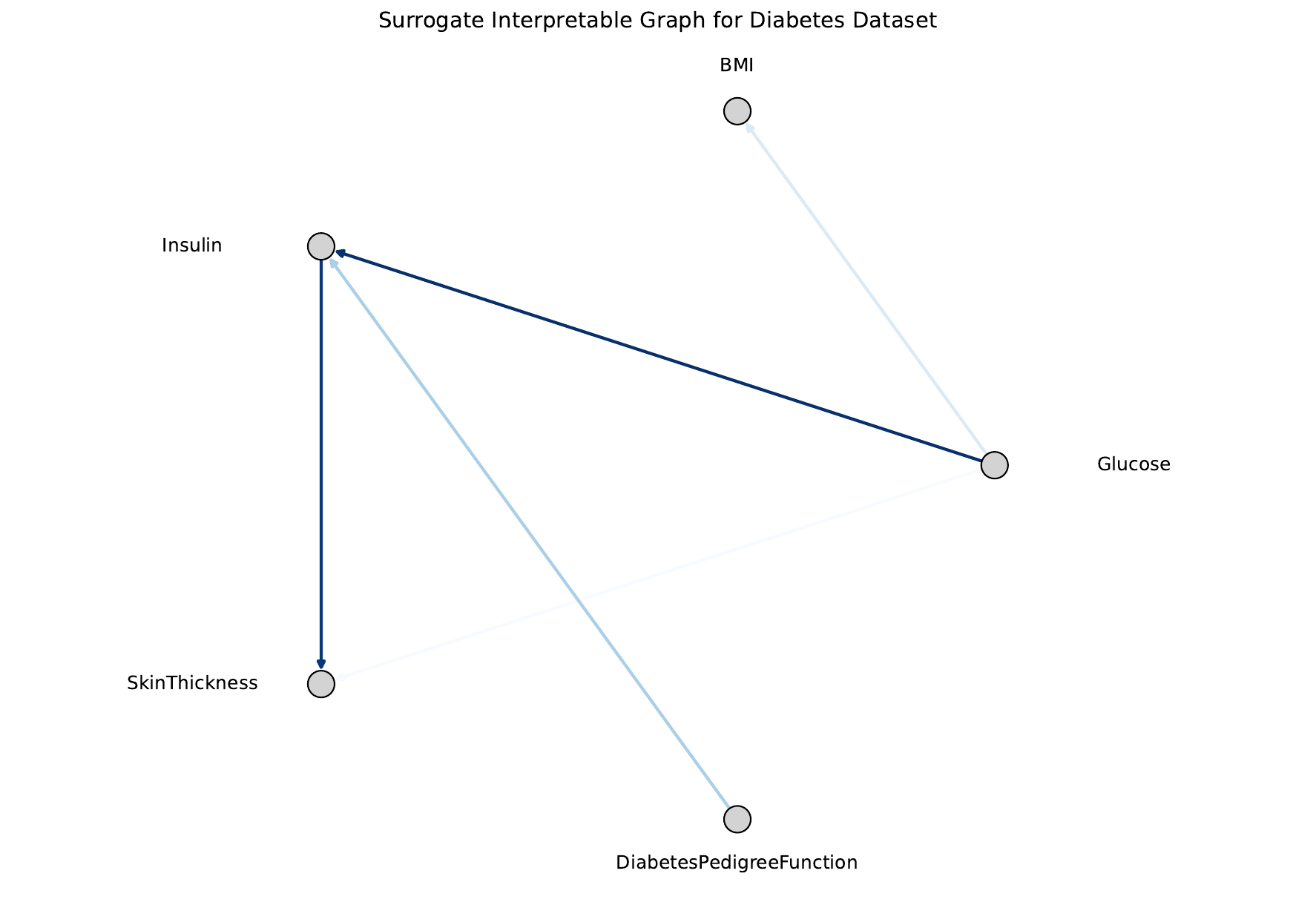}
        \caption{The optimized SIG for the diabetes dataset. The SIG has been pruned to remove the nodes with minimal significance using MILP. The max number of edges for this optimized version was 5.}
        \label{fig:sig_diabetes_unoptimized}
\end{figure}
    \vspace{1em}
    % Row 2

\begin{figure}   
        \includegraphics[width=\linewidth]{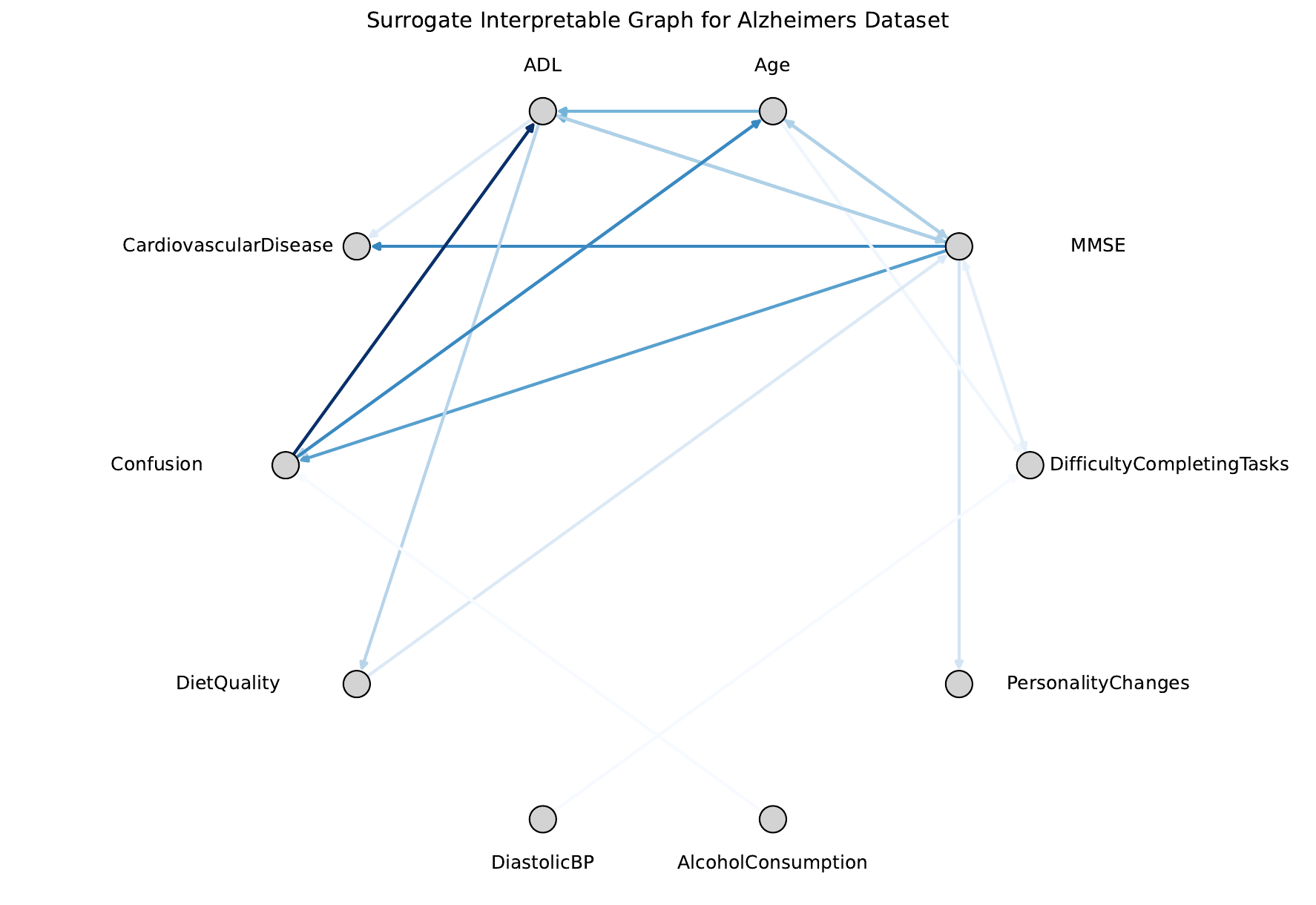}
        \caption{The optimized surrogate interpretable graph (SIG) for the Alzheimer dataset. The SIG has been pruned to remove the nodes with minimal significance using MILP. The maximum number of edges for this optimized version was 20.}
        \label{fig:sig_Alzheimer_optimized}
\end{figure}

\subsection{Heart Dataset}
The heart disease~\cite{heart_disease_45} dataset contains 14 features and 303 rows. or a detailed description of the features, see Appendix:~\ref{appendix:heart_data}. The trained RF model was used to create an unoptimized graph representation, which contains features with minimal interaction, hindering user comprehension. To optimize and prune the graph, the MILP was used to retain the most important features with the highest interaction and remove the features with minimal interaction and contribution. This resulted in a Surrogate Interpretable Graph (SIG) for the heart dataset (Figure:~\ref{fig:sig_heart_optimized}). 
% \begin{figure}
%     \centering
%     \includegraphics[width=0.8\linewidth]{figures/new_heart_optimized_surrogate_graph.pdf}
%     \caption{The optimized surrogate interpretable graph (SIG) for the heart dataset. The SIG has been pruned to remove the nodes with minimal significance using MILP. The max number of edges for this optimized version was 15.}
%     \label{fig:sig_heart_optimized}
% \end{figure}
From the MILP-optimized graph, we can see that the features with the most interactions are \textit{f1: chest\_pain\_type, f2: age, f3: max\_heart\_rate\_achieved, f4: fasting\_blood\_sugar, f5: 
number\_major\_vessels, f6: sex, f7: st\_slope, f8: st\_depression, f9: thalassemia, f10: resting\_electrocardiogram}. The SIG was pruned to remove nodes with minimal significance using MILP, resulting in a maximum number of edges of 15. The feature usage per DFI table revealed 10 important Decision Factors (DFIs) for decision making, with f1 occurring in 5 DFIs, f2 in 3 DFIs, f3 in 6 DFIs, f4 in 6 DFIs, f5 in 3 DFIs, f6 in 4 DFIs, f7 in 3 DFIs, f8 in 3 DFIs, f9 in 7 DFIs, and f10 in 2 DFIs. The most important interacting feature is f9, and the features f8 and f9 originate from the feature interaction for decision making (Table:~\ref{tab:adjacency_matrix_sig_heart}).

\begin{minipage}[t]{0.49\textwidth} % Increased width from 0.1\textwidth to 0.45\textwidth
\centering
\captionof{table}{Feature usage per decision-feature-interaction (DFI) from SIG for the heart dataset (Figure:~\ref{fig:sig_heart_optimized}).}
\label{tab:adjacency_matrix_sig_heart}
\begin{adjustbox}{width=\textwidth}
\renewcommand{\arraystretch}{1.44}
\begin{tabular}{ccccccccccc}
% \toprule
\textbf{DFI} & \textbf{f1} & \textbf{f2} & \textbf{f3} & \textbf{f4} & \textbf{f5} & \textbf{f6} & \textbf{f7} & \textbf{f8} & \textbf{f9} & \textbf{f10} \\
% \midrule
\(i_1\) & \cellcolor{gray!30}&  & \cellcolor{gray!30}&  &  & \cellcolor{gray!30}& \cellcolor{gray!30}& \cellcolor{gray!30}&  &  \\
\(i_2\) & \cellcolor{gray!30}& \cellcolor{gray!30}& \cellcolor{gray!30}& \cellcolor{gray!30}&  &  &  &  & \cellcolor{gray!30}&  \\
\(i_3\) &  &  & \cellcolor{gray!30}& \cellcolor{gray!30}&  &  &  &  & \cellcolor{gray!30}&  \\
\(i_4\) & \cellcolor{gray!30}& \cellcolor{gray!30}&  &  &  & \cellcolor{gray!30}&  &  & \cellcolor{gray!30}&  \\
\(i_5\) & \cellcolor{gray!30}&  & \cellcolor{gray!30}&  &  &  &  &  & \cellcolor{gray!30}& \cellcolor{gray!30}\\
\(i_6\) &  &  &  &  & \cellcolor{gray!30}&  &  &  & \cellcolor{gray!30}& \cellcolor{gray!30}\\
\(i_7\) &  &  &  & \cellcolor{gray!30}& \cellcolor{gray!30}&  &  &  & \cellcolor{gray!30}&  \\
\(i_8\) & \cellcolor{gray!30}& \cellcolor{gray!30}& \cellcolor{gray!30}& \cellcolor{gray!30}&  &  &  &  & \cellcolor{gray!30}&  \\
\(i_9\) &  &  &  & \cellcolor{gray!30}& \cellcolor{gray!30}& \cellcolor{gray!30}& \cellcolor{gray!30}& \cellcolor{gray!30}&  &  \\
\(i_{10}\) &  &  & \cellcolor{gray!30}& \cellcolor{gray!30}&  & \cellcolor{gray!30}& \cellcolor{gray!30}& \cellcolor{gray!30}&  &  \\
% \bottomrule
\end{tabular}
\end{adjustbox}
\end{minipage}
\hfill
\begin{minipage}[t]{0.49\textwidth} % Increased width from 0.1\textwidth to 0.45\textwidth
\centering
\captionof{table}{Hierarchical feature interaction for the features with the most interactions for the heart dataset (from SIG figure:~\ref{fig:sig_heart_optimized}).}
\label{tab:decision_path_per_tree_sig_heart}
\begin{adjustbox}{width=\textwidth}
\begin{tabular}{cp{6.2cm}} % Match height using approximate width
% \toprule
\textbf{DFI} & \textbf{Hierarchical Feature Interaction} \\
% \midrule
\(i_1\) & f8 $\rightarrow$ f7 $\rightarrow$ f6 $\rightarrow$ f1 $\rightarrow$ f3 \\
\(i_2\) & f9 $\rightarrow$ f4 $\rightarrow$ f2 $\rightarrow$ f1 $\rightarrow$ f3 \\
\(i_3\) & f9 $\rightarrow$ f4 $\rightarrow$ f3 \\
\(i_4\) & f9 $\rightarrow$ f6 $\rightarrow$ f1 $\rightarrow$ f2 \\
\(i_5\) & f9 $\rightarrow$ f10 $\rightarrow$ f1 $\rightarrow$ f3 \\
\(i_6\) & f9 $\rightarrow$ f10 $\rightarrow$ f5 \\
\(i_7\) & f9 $\rightarrow$ f4 $\rightarrow$ f5 \\
\(i_8\) & f9 $\rightarrow$ f4 $\rightarrow$ f2 $\rightarrow$ f1 $\rightarrow$ f3 \\
\(i_9\) & f8 $\rightarrow$ f7 $\rightarrow$ f6 $\rightarrow$ f4 $\rightarrow$ f5 \\
\(i_{10}\) & f8 $\rightarrow$ f7 $\rightarrow$ f6 $\rightarrow$ f4 $\rightarrow$ f3 \\
% \bottomrule
\end{tabular}
\end{adjustbox}
\end{minipage}

\subsection{Diabetes Dataset}
The dataset~\cite{smith1988using} contains 9 features and 768 rows. For a detailed description of the features, please refer to Appendix:~\ref{appendix:diabetes_data}.
The trained RF model was converted into a graph interpretation. The graph was optimized and pruned using a mixed-integer linear programming (MILP) model with 15 edges. The resulting signed graph (SIG) had 4 features and no originating features. This is an example of over-optimization of the SIG. The unoptimized graph interpretation was already interpretable and didn't require optimization, as the number of features was comparatively few compared to the other dataset we tested. The reason for over-optimization was the threshold we chose during optimization (15 edges). Only 4 features satisfied the criteria, resulting in a very simple SIG. Hence, the graph representation was again optimized, but this time the maximum number of edges was 5 (Figure:~\ref{fig:sig_diabetes_unoptimized}). 
% \begin{figure}
%     \centering
%     \includegraphics[width=0.8\linewidth]{figures/new_diabetes_optimized_surrogate_graph.pdf}
%     \caption{The optimized SIG for the diabetes dataset. The SIG has been pruned to remove the nodes with minimal significance using MILP. The max number of edges for this optimized version was 5.}
%     \label{fig:sig_diabetes_unoptimized}
% \end{figure}
The traits from f1 to f8 are abbreviated as \textit{f1: Insulin, f2: BMI, f3: Glucose, f4: Pregnancies, f5: BloodPressure, f6: Age, f7: SkinThickness, f8: DiabetesPedigreeFunction}. From the for feature usage per DFI table (Table:~\ref{tab:adjacency_matrix_diabetes}), we can interpret that there are 5 DFIs that are responsible for most of the feature-feature interactions during a decision. The feature \textit{f1} occurs in 2 DFIs, \textit{f2} occurs in 1 DFI, \textit{f3} occurs in 5 DFIs, \textit{f4} occurs in 1 DFI, \textit{f5} occurs in 1 DFI, \textit{f6} occurs in 1 DFI, \textit{f7} occurs in 1 DFI, \textit{f8} occurs in 1 DFI. From the hierarchical feature interaction table (Table:~\ref{tab:decision_path_per_tree_diabetes}), we can interpret that feature \textit{f3} is the originating feature for most of the DFIs leading to the decision.

\noindent
\begin{minipage}[t]{0.44\textwidth}
\centering
\captionof{table}{Feature usage per decision-feature-interaction (DFI) from unoptimized SIG for the diabetes dataset (Figure:~\ref{fig:sig_diabetes_unoptimized}).}
\label{tab:adjacency_matrix_diabetes}
\begin{adjustbox}{width=\textwidth}
\begin{tabular}{ccccccccc}
% \toprule
\textbf{DFI} & \textbf{f1} & \textbf{f2} & \textbf{f3} & \textbf{f4} & \textbf{f5} & \textbf{f6} & \textbf{f7} & \textbf{f8} \\
% \midrule
\(i_1\) &  &  & \cellcolor{gray!30} &  &  &  &  & \cellcolor{gray!30} \\
\(i_2\) &  & \cellcolor{gray!30} & \cellcolor{gray!30} &  &  &  &  &  \\
\(i_3\) & \cellcolor{gray!30} &  & \cellcolor{gray!30} &  &  & \cellcolor{gray!30} &  &  \\
\(i_4\) & \cellcolor{gray!30} &  & \cellcolor{gray!30} & \cellcolor{gray!30} &  &  & \cellcolor{gray!30} &  \\
\(i_5\) &  &  & \cellcolor{gray!30} &  & \cellcolor{gray!30} &  &  &  \\
% \bottomrule
\end{tabular}
\end{adjustbox}
\end{minipage}%
\hfill
\begin{minipage}[t]{0.5\textwidth}
\centering
\captionof{table}{Hierarchical Feature Interaction (From un-optimized SIG) for the features with most interactions for the diabetes dataset (Figure:~\ref{fig:sig_diabetes_unoptimized}).}
\label{tab:decision_path_per_tree_diabetes}
\begin{adjustbox}{width=\textwidth}
\renewcommand{\arraystretch}{1.33}
\begin{tabular}{cp{6.2cm}}  % Adjust column width for height alignment
% \toprule
\textbf{DFI} & \textbf{Hierarchical Feature Interaction} \\
% \midrule
\(i_1\) & f8 $\rightarrow$ f3 \\
\(i_2\) & f3 $\rightarrow$ f2 \\
\(i_3\) & f3 $\rightarrow$ f1 $\rightarrow$ f6 \\
\(i_4\) & f3 $\rightarrow$ f1 $\rightarrow$ f7 $\rightarrow$ f4 \\
\(i_5\) & f3 $\rightarrow$ f5 \\
% \bottomrule
\end{tabular}
\end{adjustbox}
\end{minipage}
\subsection{Alzheimer Dataset}
This dataset~\cite{alzheimer} contains 35 features, including but not limited to, patient ID, age, gender, BMI, smoking, alcohol consumption, physical activity, diet quality, sleep quality, systolic and diastolic blood pressure, cholesterol total, cholesterol LDL, cholesterol HDL, cholesterol triacylglycerols, MMSE, and functional assessment, memory complaints, etc.
For detailed description of the features, please refer to Appendix:~\ref{appendix:alzheimer_data}.
The dataset contains extensive health information for 2,149 patients, each with IDs from 4751 to 6900. It includes demographic details, lifestyle factors, medical history, clinical measurements, cognitive and functional assessments, symptoms, and a diagnosis of Alzheimer's disease.
The features \textit{'PatientID','Ethnicity','EducationLevel','DoctorInCharge','SystolicBP'} are removed to replicate a real-world scenario where the patient's private information is protected. Ethnicity and educational background, which may introduce unwanted bias, are also removed. The trained RF model is used to generate the graph representation. The graph is optimized \& pruned using MILP to retain the most important features with highest interaction and remove the features with minimal interaction \& contribution. This results in a SIG(Figure:~\ref{fig:sig_Alzheimer_optimized}).
% \begin{figure}
%     \centering
%     \includegraphics[width=0.8\linewidth]{figures/new_alzheimer_optimized_surrogate_graph.pdf}
%     \caption{The optimized surrogate interpretable graph (SIG) for the Alzheimer dataset. The SIG has been pruned to remove the nodes with minimal significance using MILP. The max number of edges for this optimized version was 20.}
%     \label{fig:sig_Alzheimer_optimized}
% \end{figure}
After the optimization, the features are: \textit{f1: ADL, f2: Age, f3: MMSE, f4: DifficultyCompleting Tasks, f5: PersonalityChanges, f6: AlcoholConsumption, f7: Diabetes, f8: DietQuality, f9: Confusion, f10: CardioVascularDisease}. The graph shows the most significant types of DFI (decision-feature-interaction) in the RF Classifier for predictions (Table: ~\ref{tab:adjacency_matrix_Alzheimer_heart}).The SIG provides the table for feature usage per DFI. The hierarchical feature interaction (Table:~\ref{tab:adjacency_matrix_Alzheimer_heart}) shows how features interact to make predictions. Inwards arrows indicate features' first appearance, while outwards arrows indicate features' appearance last. Two-way arrows indicate features are bi-directional.
For the Alzheimer's dataset, \textit{f1} is encountered by all the crucial DFIs. \textit{f2} is encountered by two DFIs. \textit{f3} is encountered by three DFIs. \textit{f4} is encountered by two DFIs. \textit{f5} Changes is encountered by two DFIs. \textit{f6} is encountered by two DFIs. \textit{f7} is encountered by two DFIs. \textit{f8} is encountered by two DFIs. \textit{f9} is encountered by two DFIs. \textit{f10} is encountered by two DFIs. DFI 4 has the most number of feature-interactions. The features \textit{f2}, \textit{f3}, and \textit{f4} are bi-directional. The DFIs are also less, indicating that the number of estimators could be reduced to enhance the computational complexity. For the initial optimization, the number of edges was 15, but there were no features which only had outwards arrows. This indicated that the SIG was over-optimized. To solve this, the number of edges for the optimization was increased to 20, enabling the acquisition of the features originating from the DFIs. There could be an issue with the over-optimization of the surrogate interpretable graph. This means that the graph has been over-optimized, resulting in features having only inwards arrows and outwards arrows, but no only-outwards arrows features.

\noindent
\begin{minipage}[t]{0.49\textwidth}
\centering
\captionof{table}{Feature usage per decision-feature-interaction (DFI) from SIG for the Alzheimer dataset (Figure:~\ref{fig:sig_Alzheimer_optimized}).}
\label{tab:adjacency_matrix_Alzheimer_heart}
\begin{adjustbox}{width=\textwidth}
\renewcommand{\arraystretch}{1.45}
\begin{tabular}{ccccccccccc}
% \toprule
\textbf{DFI} & \textbf{f1} & \textbf{f2} & \textbf{f3} & \textbf{f4} & \textbf{f5} & \textbf{f6} & \textbf{f7} & \textbf{f8} & \textbf{f9} & \textbf{f10} \\
% \midrule
\(i_1\) & \cellcolor{gray!30}&  &  &  &  & \cellcolor{gray!30}&  &  & \cellcolor{gray!30}& \cellcolor{gray!30}\\
\(i_2\) & \cellcolor{gray!30}&  & \cellcolor{gray!30}&  & \cellcolor{gray!30}& \cellcolor{gray!30}&  & \cellcolor{gray!30}& \cellcolor{gray!30}&  \\
\(i_3\) & \cellcolor{gray!30}& \cellcolor{gray!30}& \cellcolor{gray!30}& \cellcolor{gray!30}&  &  & \cellcolor{gray!30}&  &  & \cellcolor{gray!30}\\
\(i_4\) & \cellcolor{gray!30}& \cellcolor{gray!30}& \cellcolor{gray!30}& \cellcolor{gray!30}& \cellcolor{gray!30}&  & \cellcolor{gray!30}& \cellcolor{gray!30}&  &  \\
% \bottomrule
\end{tabular}
\end{adjustbox}
\end{minipage}
\hfill
\begin{minipage}[t]{0.49\textwidth}
\centering
\captionof{table}{Hierarchical feature interaction for the features with most interaction for the Alzheimer dataset (Figure:~\ref{fig:sig_Alzheimer_optimized}).}
\label{tab:decision_path_per_tree_alzheimer_heart}
\begin{adjustbox}{width=\textwidth}
\begin{tabular}{cp{6.2cm}}  % adjust to match height
% \toprule
\textbf{DFI} & \textbf{Hierarchical Feature Interaction} \\
% \midrule
\(i_1\) & f6 $\rightarrow$ f9 $\rightarrow$ f1 $\rightarrow$ f10 \\
\(i_2\) & f6 $\rightarrow$ f9 $\rightarrow$ f1 $\rightarrow$ f8 $\rightarrow$ f3 $\rightarrow$ f5 \\
\(i_3\) & f7 $\rightarrow$ f4 $\leftrightarrow$ f3 $\leftrightarrow$ f2 $\rightarrow$ f1 $\rightarrow$ f10 \\
\(i_4\) & f7 $\rightarrow$ f4 $\leftrightarrow$ f3 $\leftrightarrow$ f2 $\rightarrow$ f1 $\rightarrow$ f8 $\rightarrow$ f3 $\rightarrow$ f5 \\
% \bottomrule
\end{tabular}
\end{adjustbox}
\end{minipage}

\section{Comparison Between Surrogate Interpretable Graph (SIG) and TreeSHAP Interaction Values}
\label{appendix:comparison_treeshap_sig}
The comparison between SIG and TreeSHAP highlights the differences in the interpretability of feature interactions. SIG provides structural context and underlying rule logic, while TreeSHAP emphasizes meaning without considering rule logic. The graph is advantageous for interpreting features across multiple trees, condensing the logic of the forest into a human-readable representation. User-defined logic allows for selecting clustering methods, optimizing layouts, adjusting graph pruning, and adapting to dataset complexity (Table:~\ref{tab:shap_interaction_vs_sig}). TreeSHAP struggles with high-dimensional data, resulting in noisy or unreadable visualizations, and lacks intuitive validation features. Our graph provides rich maps that improve the interpretability of feature interactions.
\begin{figure}
    \centering
    \includegraphics[width=\linewidth]{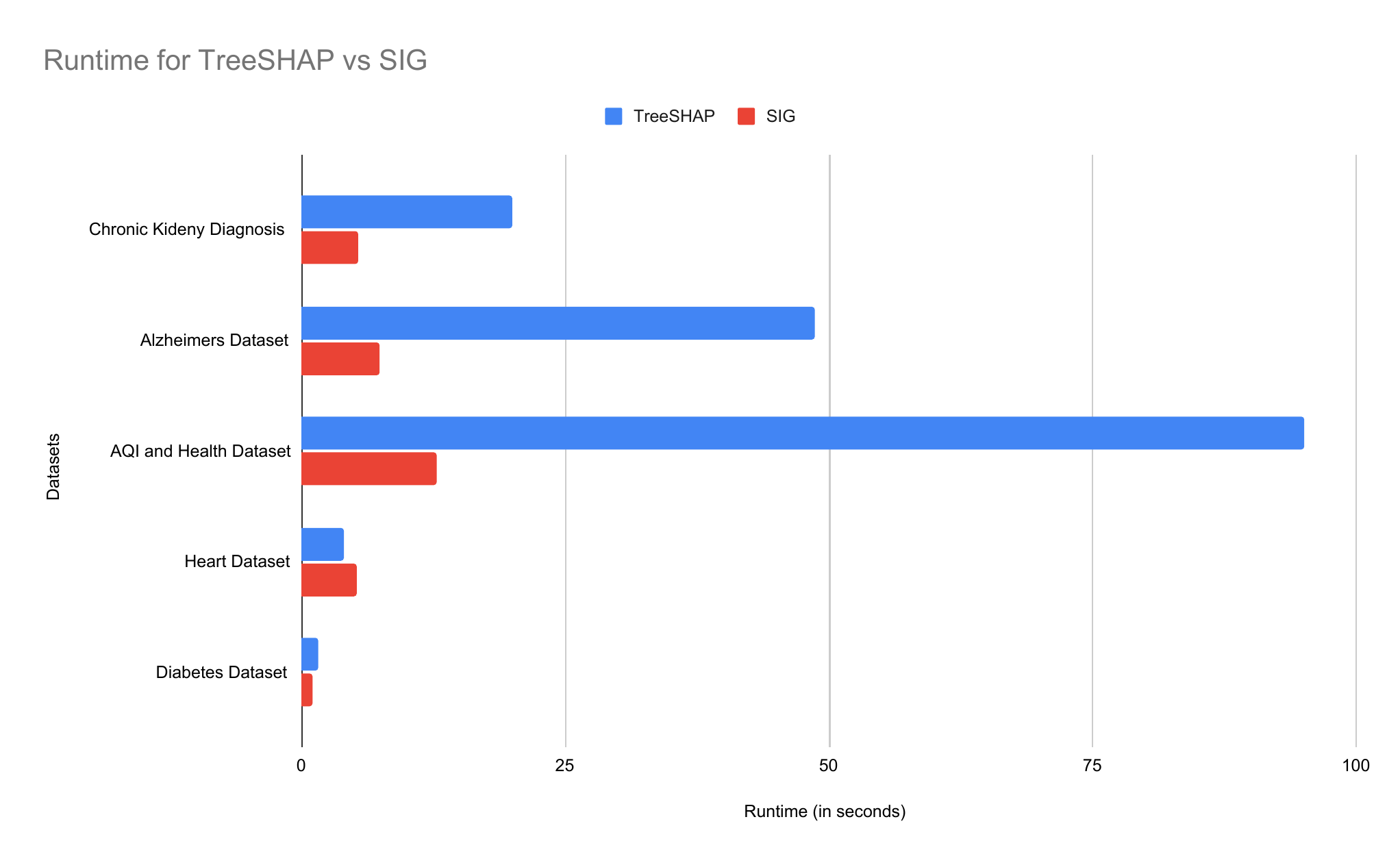}
    \caption{The comparison of runtime between SIG and TreeSHAP feature interaction for different datasets. The time complexity for SIG is higher for the basic datasets with few number of features and number of estimators in the RF classifier. The SIG gains an upper hand when the dataset size grows with respect to the number of features and data points along with the growing number of estimators.}
    \label{fig:sig_vs_treeshap_time}
\end{figure}
% \begin{table}[h!]
% \centering
% \caption{Comparison of Surrogate Interpretable Graph and TreeSHAP Interaction Values}
% \label{tab:shap_interaction_vs_sig}
% \begin{adjustbox}{width=0.9\textwidth}
% \begin{tabularx}{\textwidth}{lXX}
% \toprule
% \textbf{Method} & \textbf{Surrogate Interpretable Graph (SIG)} & \textbf{SHAP Interaction Values} \\
% \midrule
% \textbf{Transparency} & High – Shows actual rules/features the model uses. & Medium – SHAP values are additive but not always easily explainable. \\

% \textbf{Structure Awareness} & Preserves feature hierarchy and paths from trees. & Abstracts away from actual model structure. \\

% \textbf{Global Interpretability} & Excellent – Shows average co-occurrence and rule strength across forest. & Weaker – Harder to globally interpret interactions beyond pairwise SHAP plots. \\

% \textbf{Rule-Based} & Yes – Derives and clusters real decision rules. & No – Outputs contributions for a given prediction. \\

% \textbf{Visual Simplicity} & High – Graph is pruned and optimized. & SHAP plots (force, dependence) can be overwhelming. \\

% \textbf{Faithfulness to Trees} & Very faithful – Graph and rules are extracted directly from trees. & Approximate – SHAP relies on game theory assumptions. \\

% \textbf{Customizability} & High – Graph structure, clustering, rule depth, etc., can be adjusted. & Limited – SHAP has a fixed format and logic. \\
% \bottomrule
% \end{tabularx}
% \end{adjustbox}
% \end{table}
\begin{table}[h!]
\centering
\caption{Computational Efficiency Comparison of SIG with TreeSHAP}
\label{tab:sig_vs_tree_shap}
\begin{adjustbox}{width=0.9\textwidth}
\begin{tabular}{ccc}
\toprule
\textbf{Method} & \textbf{SIG} & \textbf{TreeSHAP} \\
\midrule
Time vs feature size & Sublinear if rules are sparse and compressible & Linear to exponential (via coalitions) \\

Time vs dataset size & Constant (after rule extraction) & Linear \\
Interpretability & Global, reusable across instances & Local, per-instance \\
Memory use & Efficient with rule merging & Expensive with large \(T\) and \(f\) \\
\bottomrule
\end{tabular}
\end{adjustbox}
\end{table}
The study found that the computation time for SIG and TreeSHAP is similar for small datasets such as Heart and Diabetes, and sometimes faster than SIG. However, as the number of features and data points increases while maintaining the number of estimators in the RF classifier, TreeSHAP becomes computationally expensive for the AQI and Health, Chronic Kidney, and Alzheimer's datasets (Figure:~\ref{fig:sig_vs_treeshap_time}).
\section{Dataset Description}
\label{appendix:additional_data_info}
\subsection*{Heart Dataset}
\label{appendix:heart_data}
The feature descriptions are as follows:
\begin{itemize}
      \item age: Age of the patient in years.
      \item sex: Biological sex of the patient (1 = male; 0 = female).
      \item chest\_pain\_type: Type of chest pain experienced (e.g., typical angina, atypical angina, non-anginal pain, asymptomatic).
      \item resting\_blood\_pressure: Resting blood pressure (in mm Hg) on admission to the hospital.
      \item cholesterol: Serum cholesterol in mg/dl.
      \item fasting\_blood\_sugar: Whether the patient’s fasting blood sugar > 120 mg/dl (1 = true; 0 = false).
      \item resting\_electrocardiogram: Resting electrocardiographic results (e.g., normal, ST-T wave abnormality, left ventricular hypertrophy).
      \item max\_heart\_rate\_achieved: Maximum heart rate achieved during exercise testing.
      \item exercise\_induced\_angina: Whether exercise induced angina (1 = yes; 0 = no).
      \item st\_depression: ST depression induced by exercise relative to rest.
      \item st\_slope: Slope of the peak exercise ST segment (e.g., upsloping, flat, downsloping).
      \item num\_major\_vessels: Number of major vessels (0–3) colored by fluoroscopy.
      \item thalassemia: Thalassemia blood disorder status (e.g., normal, fixed defect, reversible defect).
      \item target: Diagnosis of heart disease (0 = absence, 1 = presence of heart disease).
\end{itemize}

\subsection*{Diabetes Dataset}
\label{appendix:diabetes_data}
The feature descriptions are as follows:
\begin{itemize}
  \item Pregnancies – Number of times the patient has been pregnant.
  \item Glucose – Plasma glucose concentration after a 2-hour oral glucose tolerance test.
  \item BloodPressure – Diastolic blood pressure (in mm Hg).
  \item SkinThickness – Triceps skinfold thickness (in mm), a measure of body fat.
  \item Insulin – 2-hour serum insulin (in mu U/ml).
  \item BMI – Body mass index, calculated as weight in kg divided by height in m squared.
  \item DiabetesPedigreeFunction – A function that scores the likelihood of diabetes based on family history.
  \item Age – Age of the patient in years.
  \item Outcome – Class label indicating diabetes status (1 = diabetes, 0 = no diabetes).
\end{itemize}

\subsection*{Alzheimer Dataset}
\label{appendix:alzheimer_data}
The feature descriptions are as follows:
\begin{itemize}
    \item Age: The age of the patients ranges from 60 to 90 years.
    \item Gender: Gender of the patients, where 0 represents Male and 1 represents Female.
    \item BMI: Body Mass Index of the patients, ranging from 15 to 40.
    \item Smoking: Smoking status, where 0 indicates No and 1 indicates Yes.
    \item AlcoholConsumption: Weekly alcohol consumption in units, ranging from 0 to 20.
    \item PhysicalActivity: Weekly physical activity in hours, ranging from 0 to 10.
    \item DietQuality: Diet quality score, ranging from 0 to 10.
    \item SleepQuality: Sleep quality score, ranging from 4 to 10.
    \item FamilyHistoryAlzheimers: Family history of Alzheimer's Disease, where 0 indicates No and 1 indicates Yes.
    \item CardiovascularDisease: Presence of cardiovascular disease, where 0 indicates No and 1 indicates Yes.
    \item Diabetes: Presence of diabetes, where 0 indicates No and 1 indicates Yes.
    \item Depression: Presence of depression, where 0 indicates No and 1 indicates Yes.
    \item HeadInjury: History of head injury, where 0 indicates No and 1 indicates Yes.
    \item Hypertension: Presence of hypertension, where 0 indicates No and 1 indicates Yes.
    \item SystolicBP: Systolic blood pressure, ranging from 90 to 180 mmHg.
    \item DiastolicBP: Diastolic blood pressure, ranging from 60 to 120 mmHg.
    \item CholesterolTotal: Total cholesterol levels, ranging from 150 to 300 mg/dL.
    \item CholesterolLDL: Low-density lipoprotein cholesterol levels, ranging from 50 to 200 mg/dL.
    \item CholesterolHDL: High-density lipoprotein cholesterol levels, ranging from 20 to 100 mg/dL.
    \item CholesterolTriglycerides: Triglycerides levels, ranging from 50 to 400 mg/dL.
    \item MMSE: Mini-Mental State Examination score, ranging from 0 to 30. Lower scores indicate cognitive impairment.
    \item FunctionalAssessment: Functional assessment score, ranging from 0 to 10. Lower scores indicate greater impairment.
    \item MemoryComplaints: Presence of memory complaints, where 0 indicates No and 1 indicates Yes.
    \item BehavioralProblems: Presence of behavioral problems, where 0 indicates No and 1 indicates Yes.
    \item ADL: Activities of Daily Living score, ranging from 0 to 10. Lower scores indicate greater impairment.
    \item Confusion: Presence of confusion, where 0 indicates No and 1 indicates Yes.
    \item Disorientation: Presence of disorientation, where 0 indicates No and 1 indicates Yes.
    \item PersonalityChanges: Presence of personality changes, where 0 indicates No and 1 indicates Yes.
    \item DifficultyCompletingTasks: Presence of difficulty completing tasks, where 0 indicates No and 1 indicates Yes.
    \item Forgetfulness: Presence of forgetfulness, where 0 indicates No and 1 indicates Yes.
    \item Diagnosis: Diagnosis status for Alzheimer's Disease, where 0 indicates No and 1 indicates Yes.
\end{itemize}

\subsection*{Chronic Kidney Disease Dataset}
\label{appendix:kidney_data}
This dataset~\cite{kidney} contains 54 features and detailed health information for 1,659 patients. It Each patient is identified by a unique Patient ID, and the data includes a confidential column indicating the doctor in charge. Features like 'PatientID', 'Ethnicity', 'Socioeconomic Status', 'Education Level', and 'Doctor In Charge' are removed to replicate a real-world scenario where patient privacy is protected and treated as confidential. Ethnicity and education level can contribute to unwanted bias in the trained model. The feature descriptions are as follows:
\begin{itemize}
    \item Age: The age of the patients ranges from 20 to 90 years.
    \item Gender: Gender of the patients, where 0 represents Male and 1 represents Female.
    \item BMI: Body Mass Index of the patients, ranging from 15 to 40.
    \item Smoking: Smoking status, where 0 indicates No and 1 indicates Yes.
    \item AlcoholConsumption: Weekly alcohol consumption in units, ranging from 0 to 20.
    \item PhysicalActivity: Weekly physical activity in hours, ranging from 0 to 10.
    \item DietQuality: Diet quality score, ranging from 0 to 10.
    \item SleepQuality: Sleep quality score, ranging from 4 to 10.
    \item FamilyHistoryKidneyDisease: Family history of kidney disease, where 0 indicates No and 1 indicates Yes.
    \item FamilyHistoryHypertension: Family history of hypertension, where 0 indicates No and 1 indicates Yes.
    \item FamilyHistoryDiabetes: Family history of diabetes, where 0 indicates No and 1 indicates Yes.
    \item PreviousAcuteKidneyInjury: History of previous acute kidney injury, where 0 indicates No and 1 indicates Yes.
    \item UrinaryTractInfections: History of urinary tract infections, where 0 indicates No and 1 indicates Yes.
    \item SystolicBP: Systolic blood pressure, ranging from 90 to 180 mmHg.
    \item DiastolicBP: Diastolic blood pressure, ranging from 60 to 120 mmHg.
    \item FastingBloodSugar: Fasting blood sugar levels, ranging from 70 to 200 mg/dL.
    \item HbA1c: Hemoglobin A1c levels, ranging from 4.0% to 10.0%.
    \item SerumCreatinine: Serum creatinine levels, ranging from 0.5 to 5.0 mg/dL.
    \item BUNLevels: Blood Urea Nitrogen levels, ranging from 5 to 50 mg/dL.
    \item GFR: Glomerular Filtration Rate, ranging from 15 to 120 mL/min/1.73 m².
    \item ProteinInUrine: Protein levels in urine, ranging from 0 to 5 g/day.
    \item ACR: Albumin-to-Creatinine Ratio, ranging from 0 to 300 mg/g.
    \item SerumElectrolytesSodium: Serum sodium levels, ranging from 135 to 145 mEq/L.
    \item SerumElectrolytesPotassium: Serum potassium levels, ranging from 3.5 to 5.5 mEq/L.
    \item SerumElectrolytesCalcium: Serum calcium levels, ranging from 8.5 to 10.5 mg/dL.
    \item SerumElectrolytesPhosphorus: Serum phosphorus levels, ranging from 2.5 to 4.5 mg/dL.
    \item HemoglobinLevels: Hemoglobin levels, ranging from 10 to 18 g/dL.
    \item CholesterolTotal: Total cholesterol levels, ranging from 150 to 300 mg/dL.
    \item CholesterolLDL: Low-density lipoprotein cholesterol levels, ranging from 50 to 200 mg/dL.
    \item CholesterolHDL: High-density lipoprotein cholesterol levels, ranging from 20 to 100 mg/dL.
    \item CholesterolTriglycerides: Triglycerides levels, ranging from 50 to 400 mg/dL.
    \item ACEInhibitors: Use of ACE inhibitors, where 0 indicates No and 1 indicates Yes.
    \item Diuretics: Use of diuretics, where 0 indicates No and 1 indicates Yes.
    \item NSAIDsUse: Frequency of NSAIDs use, ranging from 0 to 10 times per week.
    \item Statins: Use of statins, where 0 indicates No and 1 indicates Yes.
    \item AntidiabeticMedications: Use of antidiabetic medications, where 0 indicates No and 1 indicates Yes.
    \item Edema: Presence of edema, where 0 indicates No and 1 indicates Yes.
    \item FatigueLevels: Fatigue levels, ranging from 0 to 10.
    \item NauseaVomiting: Frequency of nausea and vomiting, ranging from 0 to 7 times per week.
    \item MuscleCramps: Frequency of muscle cramps, ranging from 0 to 7 times per week.
    \item Itching: Itching severity, ranging from 0 to 10.
    \item QualityOfLifeScore: Quality of life score, ranging from 0 to 100.
    \item HeavyMetalsExposure: Exposure to heavy metals, where 0 indicates No and 1 indicates Yes.
    \item OccupationalExposureChemicals: Occupational exposure to harmful chemicals, where 0 indicates No and 1 indicates Yes.
    \item WaterQuality: Quality of water, where 0 indicates Good and 1 indicates Poor.
    \item MedicalCheckupsFrequency: Frequency of medical check-ups per year, ranging from 0 to 4.
    \item MedicationAdherence: Medication adherence score, ranging from 0 to 10.
    \item HealthLiteracy: Health literacy score, ranging from 0 to 10.
    \item Diagnosis: Diagnosis status for Chronic Kidney Disease, where 0 indicates No and 1 indicates Yes.
\end{itemize}

\subsection*{AQI and Health Dataset}
\label{appendix:aqi_data}
This dataset~\cite{aqi} contains 15 features. The feature descriptions are as follows:
\begin{itemize}
    \item AQI: Air Quality Index, a measure of how polluted the air currently is or how polluted it is forecast to become.
    \item PM10: Concentration of particulate matter less than 10 micrometers in diameter (micro g/m³).
    \item PM2\_5: Concentration of particulate matter less than 2.5 micrometers in diameter (micro g/m³).
    \item NO2: Concentration of nitrogen dioxide (ppb).
    \item SO2: Concentration of sulfur dioxide (ppb).
    \item O3: Concentration of ozone (ppb).
    \item Temperature: Temperature in degrees Celsius (°C).
    \item Humidity: Humidity percentage (%).
    \item WindSpeed: Wind speed in meters per second (m/s).
    \item RespiratoryCases: Number of respiratory cases reported.
    \item CardiovascularCases: Number of cardiovascular cases reported.
    \item HospitalAdmissions: Number of hospital admissions reported.
    \item HealthImpactScore: A score indicating the overall health impact based on air quality and other related factors, ranging from 0 to 100.
    \item HealthImpactClass: Classification of the health impact based on the health impact score:
    \begin{enumerate} 
        \item 0: 'Very High' (HealthImpactScore >= 80)
        \item 1: 'High' (60 <= HealthImpactScore < 80)
        \item 2: 'Moderate' (40 <= HealthImpactScore < 60)
        \item 3: 'Low' (20 <= HealthImpactScore < 40)
        \item 4: 'Very Low' (HealthImpactScore < 20)
   \end{enumerate}
\end{itemize}\

\newpage

\end{document}